\documentclass[lettersize,journal]{IEEEtran}
\usepackage{amsmath,amsfonts}
\usepackage{algorithmic}
\usepackage{algorithm}
\usepackage{array}
\usepackage[caption=false,font=normalsize,labelfont=sf,textfont=sf]{subfig}
\usepackage{textcomp}
\usepackage{stfloats}
\usepackage{url}
\usepackage{verbatim}
\usepackage{graphicx}
\usepackage{cite}
\usepackage{hyperref}
\usepackage{url}

\usepackage{graphicx}
\usepackage{makecell}
\usepackage{color}
\usepackage{tabularray}
\usepackage{subcaption}
\usepackage{algorithm}
\usepackage{algorithmic}
\usepackage{tabularx}
\usepackage{tabularray}
\usepackage{amsmath}

\usepackage{wrapfig}
\usepackage[most]{tcolorbox} 

\usepackage{xspace}
\usepackage{multirow}
\newcommand{\eat}[1]{}
\newcommand{\ie}{\emph{i.e.},\xspace}
\newcommand{\eg}{\emph{e.g.},\xspace}
\newcommand{\modelname}{TrinEval\xspace}

\usepackage{xcolor}
\usepackage{lipsum}

\definecolor{myblue}{RGB}{0, 112, 192}
\definecolor{lightblue}{RGB}{234, 239, 247}
\definecolor{myred}{RGB}{255, 0, 0}
\definecolor{mygreen}{RGB}{0, 176, 80}
\definecolor{mygray}{RGB}{245, 245, 245}
\definecolor{myblack}{RGB}{70, 70, 70}
\definecolor{lightgray}{RGB}{240, 240, 240}
\definecolor{titlebg}{HTML}{FFF3CD}   
\definecolor{outerbg}{HTML}{FFF8E1}  
\definecolor{borderclr}{HTML}{FFC107}

\newtcolorbox{preouterbox}{
    enhanced,
    colback=mygray,
    colframe=black,
}

\newtcolorbox{preextractbox}{
    enhanced,
    arc=8pt,
    colback=lightblue,
    colframe=myblue,
    boxrule=1pt,
    title={\tiny{~\\}}Prompt template for extraction:\tiny{~\\},
    left=10pt,
    right=10pt,
    top=15pt,
    bottom=10pt,
    width=\linewidth,
    before skip=10pt,
    after skip=10pt
}

\newtcolorbox{preperformancebox}{
    enhanced,
    arc=8pt,
    colback=lightblue,
    colframe=myblue,
    boxrule=1pt,
    title=Prompt template for pre-investigation on LLM
Memorization w.r.t. Capability:,
    left=10pt,
    right=10pt,
    top=10pt,
    bottom=10pt,
    width=\linewidth,
    before skip=10pt,
    after skip=10pt
}

\newtcolorbox{outerbox}{
    enhanced,
    colback=mygray,
    colframe=black,
    title=Trineval Reformulation,
}

\newtcolorbox{otemplatebox}{
    enhanced,
    arc=8pt,
    colback=lightgray,
    colframe=myblack,
    boxrule=1pt,
    title=Original MCQ,
    left=10pt,
    right=10pt,
    top=20pt,
    bottom=10pt,
    width=\linewidth,
    before skip=10pt,
    after skip=10pt
}

\newtcolorbox{oexamplebox}{
    enhanced,
    arc=8pt,
    colback=lightgray,
    colframe=myblack,
    boxrule=1pt,
    title=Original MCQ Example,
    left=10pt,
    right=10pt,
    top=5pt,
    bottom=10pt,
    width=\linewidth,
    before skip=10pt,
    after skip=10pt
}

\newtcolorbox{rtemplatebox}{
    enhanced,
    arc=8pt,
    colback=outerbg,
    colframe=borderclr,
    boxrule=1pt,
    title=TrinEval MCQ,
    left=10pt,
    right=10pt,
    top=10pt,
    bottom=10pt,
    width=\linewidth,
    before skip=10pt,
    after skip=10pt
}

\newtcolorbox{rexamplebox}{
    enhanced,
    arc=8pt,
    colback=outerbg,
    colframe=borderclr,
    boxrule=1pt,
    title=TrinEval MCQ Example,
    left=10pt,
    right=10pt,
    top=10pt,
    bottom=10pt,
    width=\linewidth,
    before skip=10pt,
    after skip=10pt
}

\begin{document}

\title{Large Language Models Could Be Rote Learners}

\author{Yuyang~Xu,
        Renjun~Hu,
        Haochao~Ying$^*$,~\IEEEmembership{Member,~IEEE,}
        Jian~Wu,~\IEEEmembership{Member,~IEEE,}
        Xing~Shi,
        and~Wei~Lin
\thanks{Yuyang Xu is with the College
of Computer Science and Technology, Zhejiang University. He is also with the State Key Laboratory of Transvascular Implantation Devices and TIDRI and the Zhejiang Key Laboratory of Medical Imaging Artificial Intelligence, Hangzhou 310058, China. E-mail: xuyuyang@zju.edu.cn.}
\thanks{Renjun Hu is with the School of Data Science of Engineering, East China Normal University, Shanghai 200062, China. E-mail: rjhu@dase.ecnu.edu.cn.}
\thanks{Haochao Ying is with the State Key Laboratory of Transvascular Implantation Devices and TIDRI. He is also with the School of Public Health, Zhejiang University, Hangzhou 310058, China. E-mail: haochaoying@zju.edu.cn.}
\thanks{Jian Wu is with the Second Affiliated Hospital and Liangzhu Laboratory, Zhejiang University School of Medicine, Hangzhou 310058, China. He is also with the State Key Laboratory of Transvascular Implantation Devices and TIDRI, and the Zhejiang Key Laboratory of Medical Imaging Artificial Intelligence, Hangzhou 310058, China. E-mail: wujian2000@zju.edu.cn.}
\thanks{Xing Shi and Wei Lin are with the Alibaba Group, China. E-mail: \{shubao.sx, weilin.lw\}@alibaba-inc.com.}
\thanks{($^*$Corresponding author: Haochao Ying.)}
}

\markboth{Journal of \LaTeX\ Class Files,~Vol.~14, No.~8, August~2021}%
{Shell \MakeLowercase{\textit{et al.}}: A Sample Article Using IEEEtran.cls for IEEE Journals}


\maketitle

\begin{abstract}
Benchmark-based evaluation, \eg multiple-choice questions (MCQs) and open-ended questions (OEQs), is widely used for evaluating Large Language Models (LLMs), yet their reliability is undermined by benchmark contamination. When pre-exposed to testing benchmarks during training, less capable LLMs have been found to achieve inflated performance, thereby yielding erroneous results. In this study, we reframe contamination as an inherent aspect of learning and seek to disentangle and expose genuine capability acquisition from superficial memorization in LLM evaluation. Following this, firstly, by analyzing model performance under different memorization conditions of MCQs, we uncover a counterintuitive trend: LLMs perform worse on memorized benchmarks than on non-memorized ones, indicating the coexistence of two learning phenomena, \ie rote memorization and genuine capability learning. To disentangle them, we propose \textbf{\modelname}, a novel evaluation framework that reformulates MCQs into an alternative knowledge-centric trinity format, reducing memorization while preserving inherent knowledge, enabling the evaluation of genuine capability in the presence of memorization. Extensive experiments validate the effectiveness and robustness of \modelname in reformulating benchmarks, and the evaluation results further reveal that mainstream LLMs rely on rote memorization for an average of 19.6\% of knowledge points across the MMLU and the GSM8K dataset. 
\end{abstract}


\section{Introduction}\label{sec:intro}



The rapid advancement of Large Language Models (LLMs), driven primarily by large-scale pre-training on massive datasets, has endowed these models with 
remarkable proficiency across diverse tasks~\cite{ouyang2022training, openai2024hello, touvron2023llama}. As LLMs continue to improve, evaluating their genuine capacities has emerged as one of the fundamental challenges, necessitating proper methodologies to ensure fairness and robustness~\cite{ganguli2023challenges, liu2023trustworthy}.



\begin{figure}[t]
  \centering
  \includegraphics[width=\linewidth]{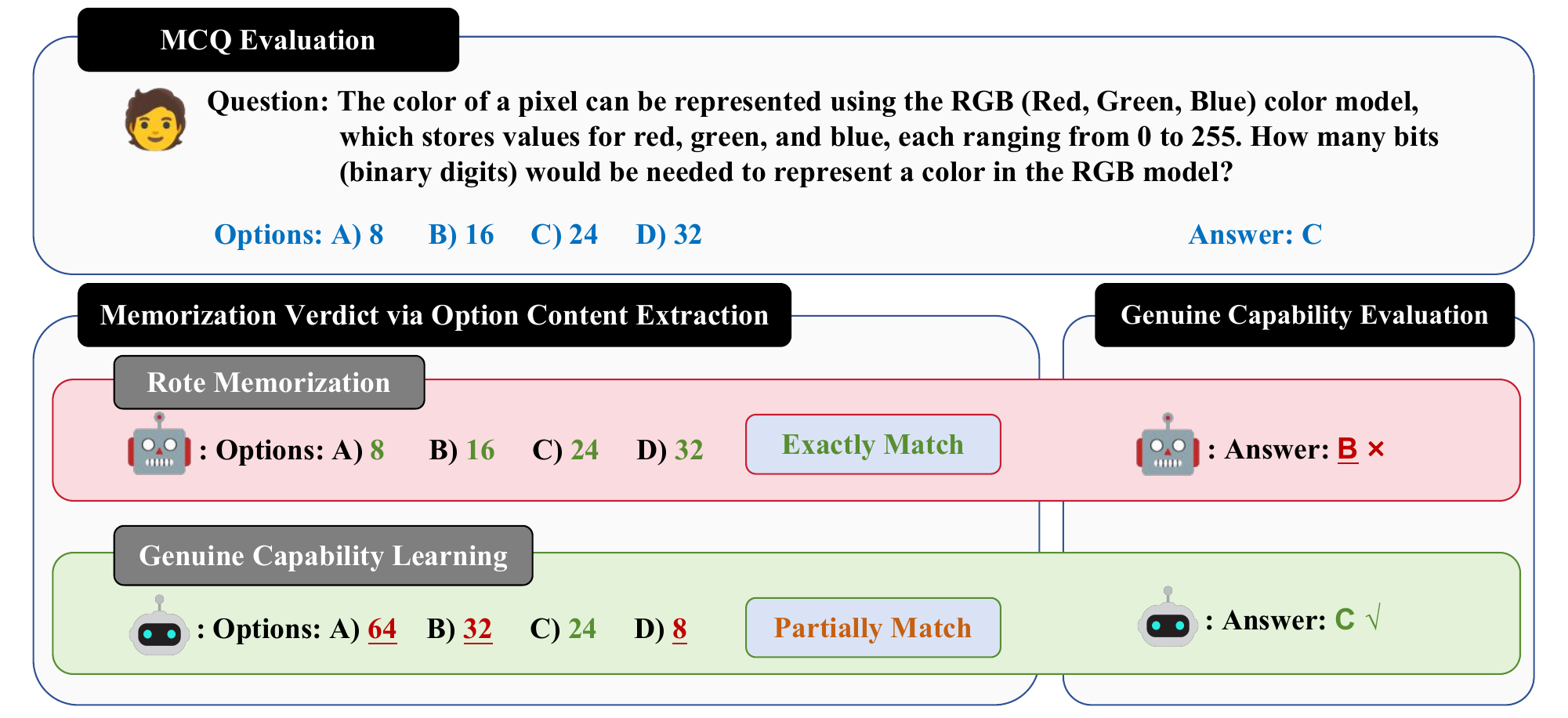}
  \caption{Benchmark-based evaluation. We observe that LLMs tend to underperform on memorized questions.}\vspace{-3ex}
  \label{fig:intro}
\end{figure}

Among the developed methods, benchmark-based evaluation has become a standard approach to evaluate LLMs. Typically, taking multiple-choice questions (MCQs) as an example, LLMs are presented with a question and a fixed set of optional choices, requiring them to select the most appropriate answer (see Fig.~\ref{fig:intro} for illustration). This format enables straightforward performance measurement through accuracy metrics and could cover a wide range of subjects. 
However, despite their widespread adoption, benchmark-based evaluation raises concerns about reliability due to benchmark contamination~\cite{li2024task, kim2024propile}, \ie test data unintentionally appears in training corpora, and LLMs may exploit memorized content rather than demonstrating genuine understanding, inflating their apparent capabilities. For instance, Zhou et al.~\cite{zhou2023don} discovers that smaller models with deliberate pre-exposure could outperform larger counterparts, thereby contradicting widely accepted scaling laws.

To mitigate this issue, Zhou et al.~\cite{zhou2023don} advocates the removal of benchmarks from pre-training corpora. 
However, this strategy conflicts with the fundamental objective of large-scale pre-training, which aims to maximize model performance by exposing LLMs to as much data as possible. 
From a broader perspective, human learning also involves problem-solving through practicing on similar questions, \eg exam preparation. While rote memorization of specific questions and answers merely leads to short-term success, repeated practice can also facilitate deeper conceptual understanding. 
Thus, rather than viewing benchmark contamination as a flaw to be eradicated, which is a nearly impossible task at scale~\cite{sainz2023nlp, bordt2024elephants}, we argue that it is an inherent aspect of learning and should be accounted for in evaluation. Therefore, this study shifts focus to \emph{evaluating LLMs in the presence of contamination, aiming to distinguish genuine capability gains from superficial memorization effects}. 
The explicit disentangling of these two learning effects remains largely unexplored in benchmark-based (MCQ-based, as an example) evaluation, yet we believe it marks a crucial step towards developing more rigorous and unbiased evaluation methodologies.



To investigate superficial memorization effects in LLM evaluation, we compare model performance under different memorization conditions. Inspired by membership inference attacks (MIA)~\cite{carlini2022quantifying,carlini2021extracting}, 
we consider the intuitive manifestation of superficial memorization is the LLM’s ability to verbatim reproduce content, \eg extracting options of MCQs in our case. With the criterion, we partition the MMLU benchmark~\cite{hendrycks2020measuring}\footnote{Selected for its popularity and documented data contamination in widely used LLMs~\cite{sainz2023nlp}.} into memorized and non-memorized subsets and evaluate three open-source LLMs\footnote{Llama2-7B~\cite{touvron2023llama}, Mistral-7B-v0.2~\cite{jiang2023mistral} and Vicuna-v1.5-7B~\cite{zheng2023judging}.}. 
Surprisingly, results reveal a consistent yet counterintuitive trend: LLMs perform worse on memorized MCQs than on those not (see Fig.~\ref{fig:intro} for illustration and Fig.~\ref{fig:pre_exp} for results). This challenges the assumption that memorization inflates model performance and suggests the coexistence of two learning phenomena in LLMs: \emph{rote memorization}, where \emph{models recall content verbatim without true understanding}, and \emph{genuine capability learning}, where \emph{they internalize underlying knowledge}.
Still, our preliminary investigation needs further exploration. First, binary classification of benchmarks as either memorized/non-memorized and capable/incapable oversimplifies the nuances of memorization and capability, potentially overlooking intermediate cases. Second, our analysis could not reveal the numerical dependency between rote memorization and capability learning.

To address these challenges, we propose \textbf{\modelname}, a novel evaluation framework designed to provide a reliable measure of LLM performance by minimizing the influence of rote memorization. \modelname employs a query-based probing  (q-probing) mechanism~\cite{allen2023physics} that reformulates benchmarks into an alternative trinity format, \ie entity-attribute-context, which prevent direct content recall while preserving knowledge assessment, and exposes the genuine capability of LLMs. 
Through experiments, we demonstrate that \modelname's reformulation is knowledge-preserving, \ie maintaining testing problems' inherent knowledge requirements without introducing extra cues, and could effectively reduce memorization. Combined with a continuous superficial memorization quantification metric, \modelname reveals the in-robustness of LLMs' capability in benchmark evaluation. With MMLU, tested open-sourced LLMs only mastered 19.0\% of knowledge points while 19.6\% are memorized by rote in the meanwhile, while for OEQs in GSM8K, LLMs only mastered 17.12\% of knowledge points while 17.30\% are rote memorized, shedding light on the necessity for further optimization on the LLMs development. Our main contributions are as follows:
\begin{itemize} 
    \item We propose a novel evaluation paradigm that assesses the genuine capability of LLMs under the default condition of benchmark contamination.
    \item We introduce \modelname, a framework that gets access to the genuine knowledge stored within tokens through textual reformulation and explores LLMs' true understanding based on verbal queries.
    \item Through \modelname's memorization-capability decoupling and quantitative analysis, we reveal a counterintuitive but general finding: the more LLMs memorize, the worse their actual problem-solving capability becomes.
\end{itemize}

\section{Related Work} 

\subsection{LLM Evaluation on Benchmarks}

The rapid advancement of LLMs has driven their expansion into diverse domains, necessitating robust and fair evaluation methodologies~\cite{zheng2023judging,Hu2025TrainingAL} and platforms~\cite{2023opencompass,chiang2024chatbot}.
Among these, evaluating on MCQ benchmarks emerges as a widely adopted approach due to the ease of validation and standardized comparison across models, including MMLU~\cite{hendrycks2020measuring}, a large-scale multi-task benchmark covering 57 subjects spanning STEM, humanities, social sciences, and professional disciplines, designed to assess general knowledge and reasoning ability; AGIEval~\cite{zhong2023agieval}, a human-centric benchmark constructed from real-world public examinations such as college entrance exams and certification tests, focusing on practical knowledge application and high-level cognitive skills; C-Eval~\cite{huang2024c}, a comprehensive Chinese multi-level evaluation suite covering middle school, high school, university, and professional domains, tailored for evaluating foundation models in Chinese understanding and reasoning; and MMLU-Pro~\cite{wang2024mmlu}, an enhanced version of MMLU that filters out simplistic questions and retains complex, model-challenging items to better measure robust high-order reasoning capacity.
However, MCQ-based evaluations are not without limitations. Biases in LLM responses have been extensively studied~\cite{dai2024unifying}, revealing issues such as social biases~\cite{salewski2024context, liu2023investigating} and order sensitivity~\cite{akter2023depth}. To mitigate the latter, PriDe~\cite{zheng2023large} estimates the option positional bias after option permutation. To examine mastery of knowledge, Zhao et al.~\cite{zhao2023knowing} apply a hypothesis testing method and check rephrased-context consistency for a given question.
Benchmark contamination is arguably the most severe challenge for MCQ-based evaluations, which may result in misleadingly inflated performance~\cite{zhou2023don,li2024task}. To address this, prior studies have explored data filtering, frequently-updated test sets~\cite{white2025livebench}, and data perturbation~\cite{li2024perteval}, yet none of these can fundamentally resolve the issue.

Beyond MCQs-based benchmarks, open-ended questions (OEQs) evaluation has gained increasing attention for assessing LLMs’ generative capabilities, reasoning chains, and factual consistency in free-form responses. Representative open-ended benchmarks include GSM8K~\cite{cobbe2021training}, which focuses on multi-step mathematical reasoning with detailed chain-of-thought annotations; MATH~\cite{hendrycks2020measuring}, covering advanced algebraic, geometric, and combinatorial problems; TruthfulQA~\cite{lin2021truthfulqa}, designed to measure truthfulness and avoid falsehoods in generation; and CommonsenseQA 2.0~\cite{talmor2019commonsenseqa} for commonsense understanding in free-response settings. These benchmarks require LLMs to produce coherent, logically complete, and factually accurate answers rather than selecting from fixed options, making them more aligned with real-world applications.
Still, OEQ-based evaluation faces similar challenges from benchmark contamination. Since test samples, including full questions, reasoning steps, and reference answers, are frequently leaked into training corpora, LLMs may directly reproduce memorized rationales instead of performing genuine reasoning~\cite{li2024task, sainz2023nlp}. Also, evaluating OEQ responses introduces additional difficulties: automated metrics such as BLEU, ROUGE, or BERTScore often fail to capture logical correctness, while human evaluation is costly, inconsistent, and difficult to scale~\cite{Hu2025TrainingAL, zheng2023judging}. Recent work has attempted to mitigate these issues by perturbing question templates~\cite{li2024perteval}, using dynamic test sets~\cite{white2025livebench}, or applying model-based judges~\cite{2023opencompass, Hu2025TrainingAL}, yet few methods~\cite{jin2025disentangling} explicitly disentangle rote memorization from genuine reasoning ability in open-ended generation and provide a commonly agreed evaluation metric. As a result, reliably evaluating the true problem-solving capacity of LLMs under contamination remains an open challenge for both MCQ and OEQ benchmarks.

In this paper, instead of attempting to eliminate contamination, we evaluate LLMs under its presence, aiming to distinguish genuine capability gains from superficial memorization effects. This marks a new perspective on LLM evaluation, revealing the extent to which models truly understand concepts rather than merely memorizing data.

\subsection{LLM Memorization}


Membership inference attacks (MIA) are commonly used to determine whether a specific sample was present in a model’s training data. Initially studied in smaller models, Carlini et al.~\cite{carlini2022privacy} investigates deep learning memorization mechanisms by identifying and removing easily detectable memorized samples. In the context of LLMs, MIA has been employed to assess privacy risks, revealing that both open- and closed-source models can leak sensitive personal data when provided with related prompts~\cite{kim2024propile}.


Beyond privacy concerns, Carlini et al.~\cite{carlini2022quantifying} formally defines LLM memorization as a model’s ability to verbatim generate text sequences following a prefix prompt. Using this definition, several studies~\cite{sainz2023nlp, bordt2024elephants, carlini2021extracting} have examined mainstream LLMs, confirming widespread test data leakage across popular benchmarks.
To quantify memorization strength, researchers~\cite{shi2023detecting, zhang2024min, oren2023proving, carlini2019secret} have further explored methods such as analyzing token probability distributions in generated outputs. 
However, while these studies extensively analyze LLM memorization, few explicitly investigate how memorization influences an LLM’s problem-solving ability. In contrast, our work focuses on their interplay, presenting a more rigorous approach to fair and reliable LLM evaluation.

\section{Methodology}

\subsection{Pre-investigation of LLM Capability w.r.t. Memorization}
\label{sec:pre_exp}



Benchmark contamination often leads to inflated performance estimate. This phenomenon is commonly attributed to models memorizing specific question and answer sequence rather than demonstrating genuine problem-solving abilities. 
However, the extent to which and how memorization influences LLM performance remains unclear. 
To disentangle genuine capability acquisition from superficial memorization, we conduct a preliminary investigation into how LLMs perform under different memorization conditions. By examining model accuracy on memorized vs. non-memorized subsets, we aim to reveal the role of memorization in LLM evaluation and establish a foundation for more rigorous assessment methodologies.

In this paper, we mainly discussed the LLM Capability w.r.t. Memorization via MCQs as an example\footnote{We also discuss and validate the extension of \modelname on OEQs in Sec.~\ref{sec:mem_dist}.}. 
Formally, we define an MCQ as $x=\{x_Q, x_O, x_W\}$, where $x_Q$, $x_O$, and $x_W$ refer to the question, options, and ground-truth answer, respectively. Following the memorization definition from Carlini et al.~\cite{carlini2022quantifying}, we define an MCQ $x$ is memorized by LLM $G$ if $G$ can extract the option contents $x_O$ exactly given the question $x_Q$. 
In practice, given the commonly used next-token prediction pre-training, we incorporate meta-information (\eg benchmark name) and 5-shot examples to elicit the memorized following content \footnote{We aim to evoke the memorization of LLMs by prompting this information. For detailed reasons, please refer to Sec.~\ref{sec:exp_mem}.}, and use greedy decoding (\ie temperature fixed to 0) during extraction~\cite{bordt2024elephants, sainz2023nlp} (refer to Appendix A for the complete prompt). 
Using MMLU~\cite{hendrycks2020measuring} as the evaluation benchmark, we divide the test set MCQs into memorized and non-memorized subsets, where the memorized subset consists of 909--982 questions (accounting for 6.5\%--7.0\% of the total 14,006) depending on the tested LLMs LlaMA2-7B, Mistral-7B-v0.2, and Vicuna-v1.5-7B. The detailed statistics of questions across the memorized/non-memorized subsets are given in Tab. 1 in the Appendix A. 

\begin{figure}[t]
\centering
  \includegraphics[width=\linewidth]{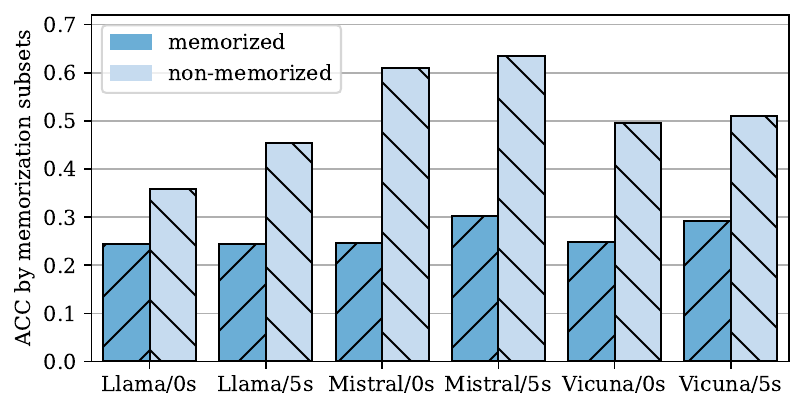} 
  \caption{Model performance on memorized and non-memorized subsets of MMLU, where `0s' and `5s' stand for zero- and five-shot prompting, respectively.}
  \label{fig:pre_exp}\vspace{-4.5ex}
\end{figure}
We then compute the accuracy (ACC) of tested LLMs by subsets as a proxy of model performance under different memorization conditions. The results of both zero- and five-shot prompting are reported in Fig.~\ref{fig:pre_exp}, from which we observe a consistent yet somehow counterintuitive trend: LLMs exhibit 47.2\% lower accuracy on average on memorized MCQs compared to non-memorized ones, regardless of LLMs and prompting techniques. This finding challenges the commonly held assumption that memorization directly improves model performance. In addition, it also implies the coexistence of two distinct learning paradigms or extremes within LLMs, which we term \textit{rote memorization} and \textit{genuine capability learning}, respectively. 
We also observe that the majority of memorized questions are those relatively simple, \ie not in MMLU-PRO~\cite{wang2024mmlu} (refer to Tab. 1 in the Appendix A), which is a commonly recognized difficult subset of MMLU
. Since MMLU-Pro’s curation prioritizes complex, model-challenging questions by excluding simplistic ones, the non-inclusion of our memorized questions in MMLU-Pro serves as objective evidence of their relative simplicity. This alignment with MMLU-Pro’s explicit filtering criterion validates our claim as reasonable, eliminating the possibility that the memorized quenstions are relatively difficult.

However, our pre-investigation has its limitations. Using ACC as the performance metric does not truly capture model capacity. Additionally, the binary classification of memorization potentially overlooks more nuanced forms of learning due to the discrete classification cutoff. We address these two issues in the following Sec.~\ref{sec:cap_eval} and Sec.~\ref{sec:mem_eval}, respectively, which then ensure a disentangled and quantitative analysis of rote memorization and capability learning.

\subsection{Measuring LLM Capability with \modelname}\label{sec:cap_eval}


First, in order to truly capture
model capacity, we present \modelname, a novel evaluation framework designed to provide a reliable measure of LLM performance by minimizing the influence of rote memorization. \modelname first preliminarily extract the knowledge-centric triplet of each question, detecting where the knowledge is stored and how to expose the knowledge through query-based probing. The framework then verify and reflect whether the trinity is qualified or not. Finally, \modelname utilizes the trinity to evaluate the genuine capability of LLMs.

To understand how LLMs store and manipulate knowledge, Allen et al.~\cite{allen2023physics} created a fictional biography dataset that enumerates various attributes (\eg, names, jobs, universities) and trained LLMs on it. 
They employed a prompt-based linear probing method to uncover correlations between the entity token embeddings and the associated attributes, revealing that where LLMs encode/store knowledge, \eg under person names or a sequence of knowledge augmentation, is crucial for robust mastery of knowledge.
This insight leads us to believe that entity tokens, which should ideally store knowledge, are the target for evaluating an LLM’s genuine capability.

However, applying this method to real-world datasets, such as MMLU, presents challenges. Unlike controlled datasets with explicitly defined attributes, real-world data includes a far broader range of possible knowledge. As a result, we cannot enumerate all potential attributes and directly apply linear probing.
To this end, we propose \modelname, a verbal query probing method that reformulates MCQs around a knowledge-centric trinity: knowledge entity, attribute, and context. \modelname is a pluggable augmentation on any MCQ benchmarks and could expose the genuine capability of LLMs by verifying whether they have correctly encoded knowledge. In contrast to Allen et al.~\cite{allen2023physics}, \modelname introduces a verbal query-based probing mechanism. It reformulates MCQs into a structured triplet format (entity–attribute–context), enabling direct assessment of knowledge encoded in entity tokens—without relying on model internals. This shift from embedding-space analysis to verbalized knowledge extraction is a conceptual and methodological advance. Generally, we instruct LLMs to extract the triplet according to the prompt and check and refine it if they are not qualified. This systematic validation is absent in most prior rephrasing methods and is empirically shown to be knowledge-preserving without introducing extraneous cues. The reformulation is completed by a two-round reflection-based prompting method, with the detailed procedure shown in Alg.~\ref{alg:reorg} and related prompts available in Fig. 2 in Appendix B. Here, we explain the elements of the trinity and how to reformulate them.



\textbf{Knowledge entity}. We suppose that if an LLM has mastered some knowledge, the key information pertinent to the knowledge should be encoded within a few subject tokens, namely, knowledge entity, to support efficient retrieval. By isolating these tokens, \modelname ensures that only the essential information is considered. 

\textbf{Attribute}. The attribute acts as a verbal probe to guide the model to focus on the specific feature or property of the knowledge entity being inquired about. This mechanism allows \modelname to isolate and assess the model’s understanding of the critical aspects of the questioning subject.

\textbf{Context}. For some questions, conditions or background contexts can significantly influence the solution approach. By explicitly including context in the evaluation process, \modelname helps the model account for relevant situational details that might otherwise be overlooked, ensuring that the answer is based on a comprehensive understanding of the problem.

Traditional rephrasing often relies on lexical variations (e.g., synonym replacement or sentence restructuring), which may still preserve memorization cues. In contrast, by extracting the core and necessary question information in this trinity format, the reformulation by \modelname is knowledge-preserving for the purposes of assessment. In the meantime, it completely destroys the original token sequence, effectively reducing the influence of memorization. We will empirically verify these properties through experiments. 

Based on the framework, we locate which token the knowledge is stored within and expose the genuine capability of LLMs with the trinity. Given an MCQ $x=(x_Q, x_O, x_W)$, it first queries a capable reformulation LLM to derive the knowledge entity $x_E$, attribute $x_A$, and context $x_C$ from the original $x$. The LLM is instructed that the triplet should be sufficient for answering the question correctly, without including the answer option itself, ensuring the integrity of the evaluation. 
In order to mimic the data contamination with in-context learning, we also present the original MCQ text in the format on the Huggingface dataset site as the ``**Fact:**'' part in the prompt, given that the potential data leakage is often caused by the data crawled on the Huggingface dataset site.
The same LLM then assesses whether the triplet contains all necessary information and no redundant information (typically, the rote-memorization), meanwhile, yields a rationale $x_L$ as reflection~\cite{shinn2024reflexion, yao2022react}. If it does, the triplet is returned as the re-formulated question. Otherwise, the reformulation model refines the extraction, taking as input $x_E$, $x_A$, $x_C$, and $x_L$, and re-evaluates the updated triplet. The details of the reorganization algorithm are shown as follows. Reflection prompts are shown in Appendix B.
\begin{algorithm}
	\renewcommand{\algorithmicrequire}{\textbf{Input:}}
	\renewcommand{\algorithmicensure}{\textbf{Output:}}
	\caption{MCQ reformulation by \modelname}\label{alg:reorg}
	\begin{algorithmic}[1]
		\REQUIRE Question $x_Q$, options $x_O$, and answer $x_W$ of an MCQ.
		\ENSURE Reformulated question $x^R_Q$.
        \STATE Preliminarily extract knowledge entity $x_E$, attribute $x_A$, and context $x_C$ based on $x_Q$, $x_O$ and $x_W$;
        \STATE Initialize $X^R_Q = {x_E, x_A, x_C}$;
        \STATE Validate the adequacy and necessity of the $x^R_Q$ and give reasons $x_L$;
        \IF{$x^R_Q$ matches the requirement}
            \STATE Return $x^R_Q$;
        \ELSE
            \STATE Re-extract $x_E'$, $x_A'$, and $x_C'$ by reflecting with $x_E, x_A, x_C$ and $x_L$;
            \STATE Update $x^R_Q = {x_E', x_A', x_C'}$;
            \STATE Validate the adequacy and necessity of the $x^R_Q$ and give reasons $x_L$;
                \IF{$x^R_Q$ matches the requirement}
                    \STATE Return $x^R_Q$;
                \ELSE
                    \STATE Discard the MCQ, return $None$;
                \ENDIF
        \ENDIF
	\end{algorithmic}  
\end{algorithm}

Finally, prompting with the extracted $x_E$, $x_A$, $x_C$, and options $x_O$ (see Fig. 5 in Appendix B for illustration), we inspect the generation probability of the next ground-truth answer token $x_W$ (\ie A/B/C/D) as the measurement:
\begin{equation}
    F_c(x, G) = p_G(x_W|x_E, x_A, x_C, x_O).
\end{equation}
\noindent As can be seen, $F_c$ metric retains the necessary knowledge-centric information while discarding unnecessary biases, especially the rote memorization of LLMs, which leads to the quantification of genuine capability of LLMs.

\subsection{Quantifying LLM Memorization}\label{sec:mem_eval}

On the other hand, in order to look into the nuanced forms of learning, we believe that a continuous metric is required for memorization evaluation. For quantifying the memorization of LLMs, prior research~\cite{shi2023detecting, zhang2024min} suggests that outlier tokens, which exhibit higher generation probabilities\footnote{As established in Shi et al.~\cite{shi2023detecting}: ``an unseen example is likely to contain a few outlier words with low probabilities under the LLM, while a seen example is less likely to have words with such low probabilities.''}, are more likely to be found in memorized samples. Experiments on WIKIMIA~\cite{shi2023detecting} also finds that this method exhibits high AUC score on seen samples. In order to avoid discrete binary classification, building on this idea, we adopt the metric that utilizes the bottom $K\%$ of token probabilities within the generated sequence as a measure of memorization (we primarily set the value of $K$ to 10 in this paper). Formally, the memorization score $F_m(\overline{x}, G)$ of LLM $G$ on text sequence $\overline{x}$ is computed as follows\footnote{In the experiment, we have also tested other similar metrics like Min-K++\%~\cite{zhang2024min}, which show consistent results (exhibit a Pearson correlation coefficient of 0.9007 (strong positive correlation)). Thus, we only use Min-K\%~\cite{shi2023detecting} as an example here}: 
\begin{equation}
    F_m(\overline{x}, G) = \frac{1}{|\mathcal{M}_K(\overline{x})|}\sum\limits_{\overline{x}_i \in \mathcal{M}_K(\overline{x})} \log p_G(\overline{x}_i|\overline{x}_{1:i-1}),
\end{equation}
\noindent where $p_G(\overline{x}_i|\overline{x}_{1:i-1})$ denotes the generation probability of token $\overline{x}_i$ by $G$ given its prefix subsequence as context, and set $\mathcal{M}_K(\overline{x})$ includes the $K\%$ of tokens with the lowest probabilities. The higher $F_m$ is, the more likely $\overline{x}$ is memorized by the LLM, \ie the least memorized content could still been extracted with a high probability. Note that our objective is not to achieve high-precision quantification of memorization. Instead, we aim to statistically demonstrate a valid quantification of memorization across vast samples for comparative analysis.

\section{Experiments}\label{sec:exp}

In this section, we conduct extensive experiments to answer the following questions:

\noindent \textbf{Q1.}~Does \modelname effectively preserve knowledge to enable accurate knowledge assessment?

\noindent \textbf{Q2.}~To what extent does \modelname mitigate memorization effects during capability evaluation?

\noindent \textbf{Q3.}~What insights does \modelname provide into the rote memorization and genuine capability learning in LLMs?

\subsection{Experiment Setup}\label{sec:exp_setup}

\textbf{Models}. 
We employ proprietary commercial LLMs for question reformulation in \modelname, \ie gpt-4o-2024-08-06(GPT)~\cite{openai2024hello}, qwen-max-2024-09-19(Qwen)~\cite{qwen2, qwen2.5}, deepseek-r1(Dpsk)~\cite{guo2025deepseek} and qwen3-max-2026-01-23(Qwen3)~\cite{yang2025qwen3}. 
For model evaluation, we utilize open-source LLMs to obtain token-level logits. Our experiments first focus on three widely adopted models: LlaMA2-7B (LlaMA)~\cite{touvron2023llama}, Mistral-7B-v0.2 (Mistral)~\cite{jiang2023mistral}, and Vicuna-v1.5-7B (Vicuna)~\cite{zheng2023judging} due to their reported data contamination evidence~\cite{sainz2023nlp, touvron2023llama} on MMLU. Then, we further extend the results to the latest larger models, such as LlaMA3.1-8B~\cite{grattafiori2024llama}, Qwen3-14B, and Qwen3-32B. These models are accessed from Hugging Face and implemented with the transformers library, allowing us to compute the log-probabilities of generated tokens for fine-grained analysis. All evaluations are conducted on a single NVIDIA A100 GPU with 40GB of memory. Default generation parameters are used, and greedy decoding is applied throughout to ensure reproducibility.


\textbf{Benchmarks}. We evaluate LLMs on the widely used MMLU benchmark~\cite{hendrycks2020measuring}, in light of growing evidence of data contamination~\cite{touvron2023llama}
in many recent LLMs on this data.
MMLU spans 57 subjects across STEM, humanities, social sciences, and others, providing a comprehensive assessment of model capabilities. 
We de-duplicate MCQs across subjects on MMLU. Also, MCQs from some subjects contain similar or identical options\footnote{\eg the options of MCQs in the subject, \textit{moral\_scenarios}, are all identical (`Wrong, Wrong', `Wrong, Not wrong', `Not wrong, Wrong' and `Not wrong, Not wrong').}. With the provided 5-shot prompt, options of MCQs from these subjects can be easily extracted, leading to a high False-positive ratio. In order to avoid the influence of the few-shot prompt on the option extraction, we eliminate MCQs in which any of the options have appeared twice in the dataset, resulting in a test set of 14,006 unique questions.
We also extend the evaluation on the widely used GSM8K benchmark~\cite{cobbe2021training}, given the determinacy of its answers and the verified data contamination~\cite{sainz2023nlp}.
GSM8K consists of high-quality grade school math problems, focusing on multi-step arithmetic reasoning to rigorously evaluate the basic mathematical problem-solving ability of LLMs. In this way, we are able to evaluate the memorization of the OEQs according to the extraction of the official chain-of-thought(COT) sequences.
We mainly select the test set of GSM8K as the evaluation benchmark. After filtering the duplicate OEQs, we obtain an OEQ set of 1,319 unique questions.

\textbf{Evaluation}. With commercial LLMs, we evaluate model performance by extracting predicted answers with regular expressions on MCQs. For open-source models, we access the output probability of the first generated token (\ie corresponding to multiple-choice option IDs A/B/C/D) to compute quantitative performance metric for MCQs. For OEQs, we compute the perplexity of the answer string as the capability metric, only available with the open-source LLMs.


\subsection{Q1. Is \modelname Knowledge-preserving?}

We begin by verify whether reformulation of \modelname preserves essential knowledge, a prerequisite for reliable knowledge assessment. Specifically, our goal is to ensure that (1) the reformulation does not omit critical information that would cause originally correctly answered questions to be answered incorrectly, and (2) it does not introduce anomalous or unexpected content that could artificially inflate performance. We verified the above two objectives through both automatic analysis based on LLMs' results and manual annotations, ensuring the robustness of \modelname.


\subsubsection*{Q1.1 Automatic analysis}

\begin{figure}[t]
    \centering
    \includegraphics[width=\linewidth]{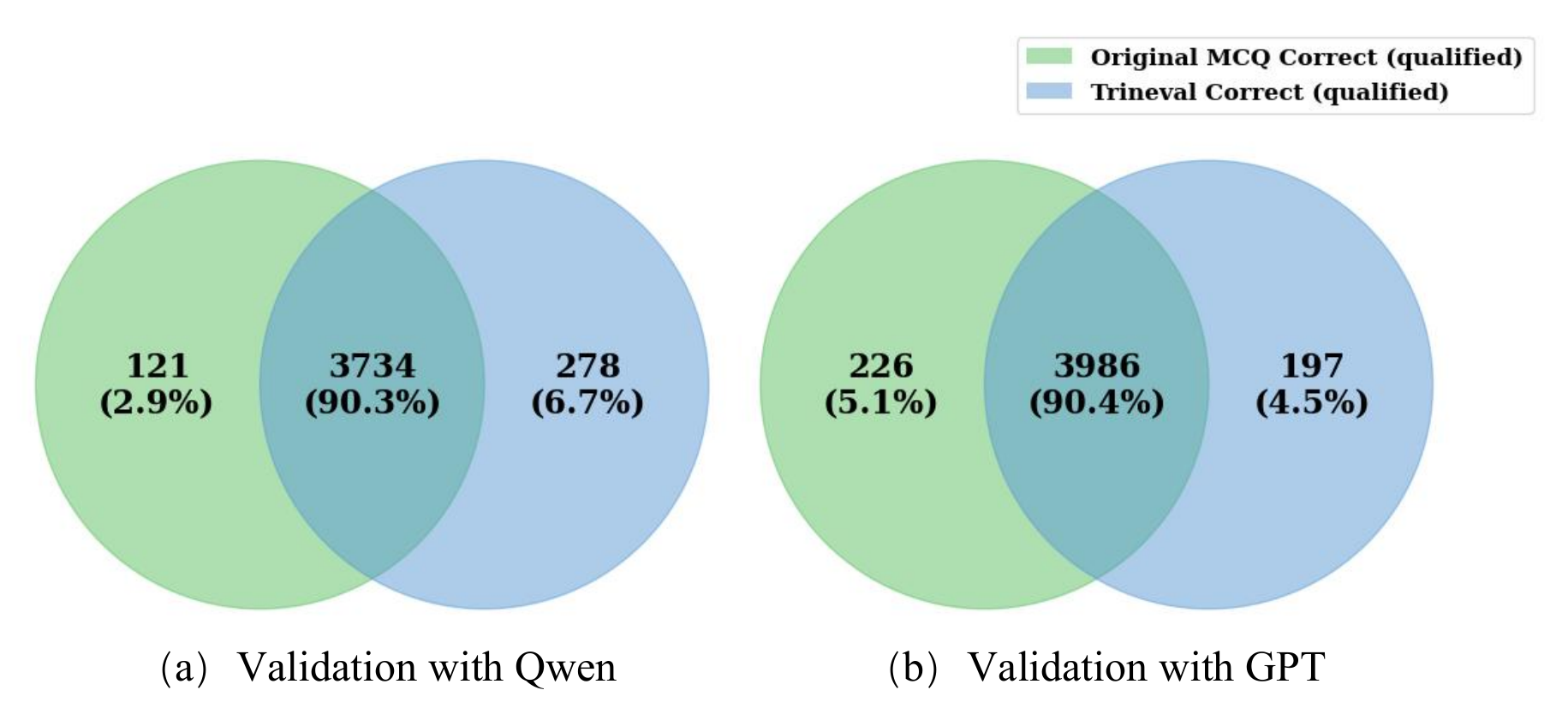} 
    \caption{
    Knowledge-preserving validation of \modelname. 
    The green, blue, and overlapping regions represent the sets of MCQs correctly answered in the original format, \modelname format, and both formats, respectively. Best viewed in color.
    }\label{fig:validate_adequacy}\vspace{-2ex}
\end{figure}

Following the complete \modelname reformulation pipeline, we obtain 4,343 qualified MCQs with corresponding triplets using Qwen, and 4,645 MCQs with associated triplets using GPT. We then prompt both Qwen and GPT to answer their respective MCQs in both the original and reformulated triplet-based formats. Example MCQs and the questioning prompt templates in both formats are provided in Fig. 5 in Appendix B; results are summarized in Fig.~\ref{fig:validate_adequacy}. Here, we primarily employ robust proprietary LLMs to evaluate the knowledge retention capability of the \modelname framework. This approach effectively mitigates potential interference from the models' memorization effects, thereby ensuring accurate assessment of their intrinsic knowledge processing abilities and preventing erroneous conclusions that might arise from conflating memorization with genuine comprehension.


Among the qualified MCQs reformulated by Qwen and GPT, 4,133 and 4,409 are answered correctly in at least one of the two formats. Notably, the majority of these are correctly answered in both original and \modelname formats, accounting for 90.3\% with Qwen and 90.4\% with GPT, indicating strong consistency in terms of problem-solving across formats. Only 121 questions for Qwen (2.9\%) and 226 for GPT (5.1\%) are answered correctly exclusively in the original format, suggesting the \modelname reformulation does not omit essential information. Conversely, 278 questions for Qwen (6.7\%) and 197 for GPT (4.5\%) are answered correctly only in the reformulated format, which is comparable in general to the numbers for the original format alone. This further suggests that the reformulation does not introduce extra information that might artificially enhance model performance. 

\subsubsection*{Q1.2 Manual annotation}

Given the potential risk for the LLM-based Knowledge Preserving evaluation (line 4 and line 10 in \ref{alg:reorg}), the proprietary LLMs might still be able to answer the MCQs without sufficient knowledge. Since this is still prompting LLMs who have been trained on these datasets and ``know'' the original content, human annotation is also applied. The annotation of each MCQ encompasses three subtasks: (1) answering the question in the re-organized form, (2) answering the question in the original form, and (3) verifying if the reformulation is Knowledge Preserving or not.

\begin{table}[ht]
\centering
\begin{tabular}{l|cc}
\hline
\textbf{Metric} & \textbf{Qwen} & \textbf{GPT} \\ \hline
\textbf{\modelname correct only} & 0.0\% & 0.0\% \\
\textbf{both correct}            & 96.296\% & 96.667\% \\
\textbf{original correct only}   & 3.704\% & 3.333\% \\
\textbf{K.P. score}              & 4.101 & 4.369 \\ \hline
\end{tabular}
\caption{Result of Human Annotation on LLM-based knowledge preserving (K.P.) evaluation in \modelname}\label{tab:human_annot}
\end{table}

For efficient annotation, we implemented a stratified sampling procedure by selecting one MCQ per subject from all 56 MMLU subjects (as a temporary compromise for limited time, which will be expanded later) under both Qwen and GPT reorganization paradigms. This yielded 112 representative questions (2 systems × 56 subjects) for evaluation. Three human annotators independently performed dual-form assessments through: (1) Direct question answering with the reformed format first and the original format; (2) Knowledge Preservation (K.P.) scoring across two dimensions: i. Knowledge adequacy (sufficiency for accurate response), ii. removal of redundant content using a 5-point scale (1=unsatisfactory; 2=major information is missed or unnecessary information is incorporated, but part is still acceptable, 3=need to take some time to understand, but can still solve the MCQ, 4=an element properly belonging to one triplet component appears in another, but does not impact MCQ solving; 5=optimal). We use a continuous rather than a binary metric to mitigate the cognitive difference in the threshold between the annotators. Inter-rater reliability was ensured through consensus-building discussions prior to formal annotation. Final scores were aggregated using averaged values to further mitigate individual annotator bias. The results are shown in Tab.~\ref{tab:human_annot} below.


Our analysis reveals that over 95\% of correctly answered MCQs maintained consistency across both original and paraphrased formats. Furthermore, human annotators rated our paraphrased questions mean K.P. score exceeding 4.0 (on a 5-point scale), which means that the reformulated MCQs only somehow influence the readability of humans but do not impact the solvability of the original format. This provides empirical validation that our proposed \modelname methodology effectively preserves necessary knowledge elements from original MCQ formulations, while the influence of the LLM memorization during the evaluation is rather limited.

Experimental results under human annotation also reveal that Qwen underperforms GPT in key metrics, particularly in processing long-context texts, where it occasionally omits background information (evidenced by excessive "N/A" assignments in the Context fields). This capability gap is further reflected in MCQ annotations: there are only MCQs that are merely correctly answered with the original format, except for the correct MCQs with both formats. Collectively, these findings provide strong evidence that \modelname preserves the core knowledge necessary for answering, satisfying the requirements for reliable capability evaluation.

\subsection{Q2. Can \modelname Reduce Memorization?}\label{sec:exp_mem}

\begin{figure*}[ht]
    \centering
    \subfloat[Evoc. with LlaMA2 and dev-fsp.\label{fig:mempert_subfig4}]{
        \includegraphics[width=0.325\textwidth]{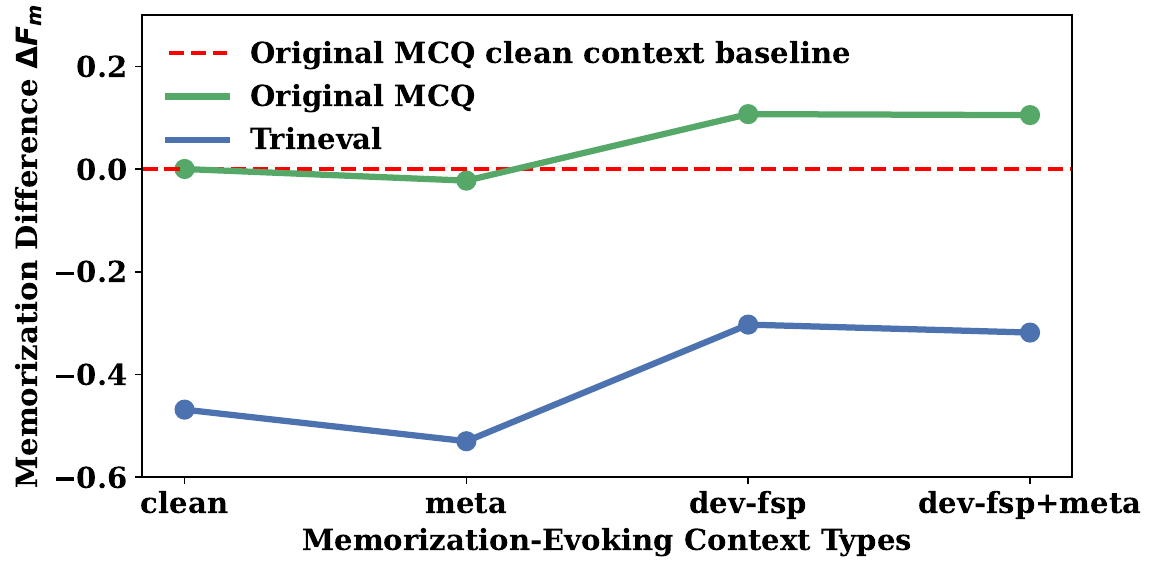} 
        }
    \subfloat[Evoc. with Mistral and dev-fsp.\label{fig:mempert_subfig2}]{
        \includegraphics[width=0.325\textwidth]{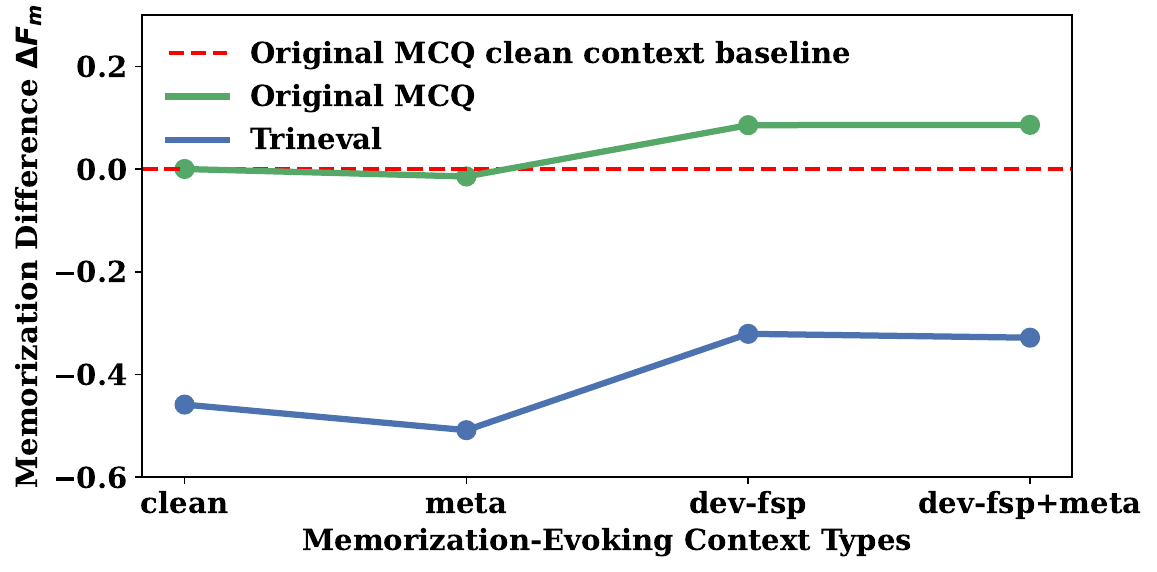} 
        }
    \subfloat[Evoc. with Mistral and seq-fsp.\label{fig:mempert_subfig5}]{
        \includegraphics[width=0.325\textwidth]{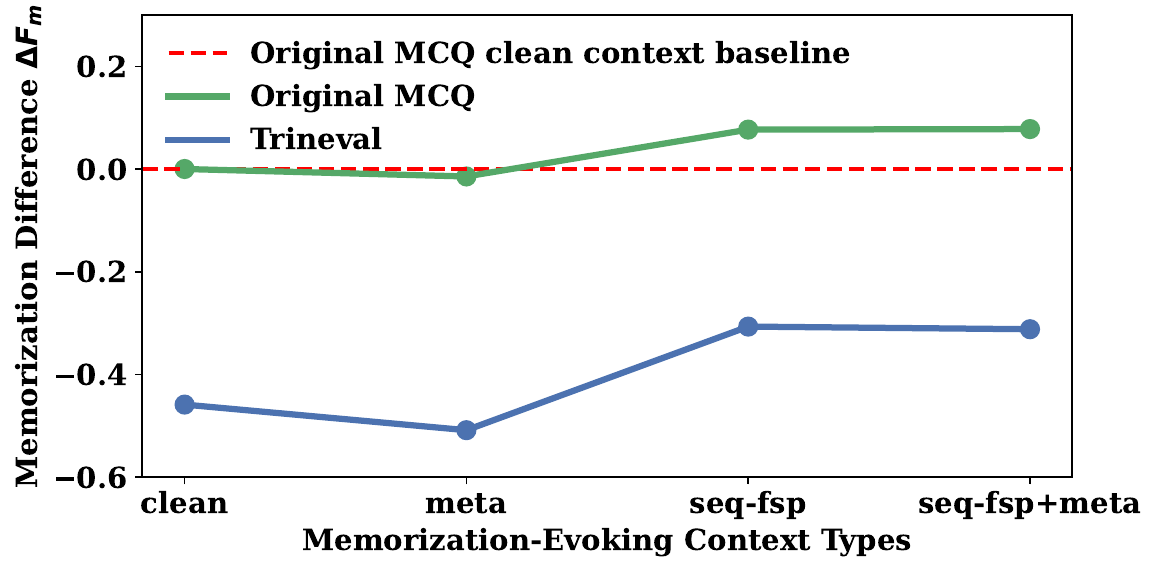} 
        }

    \vfill
    
    \subfloat[Evoc. with LlaMA2 and seq-fsp.]{
        \includegraphics[width=0.325\textwidth]{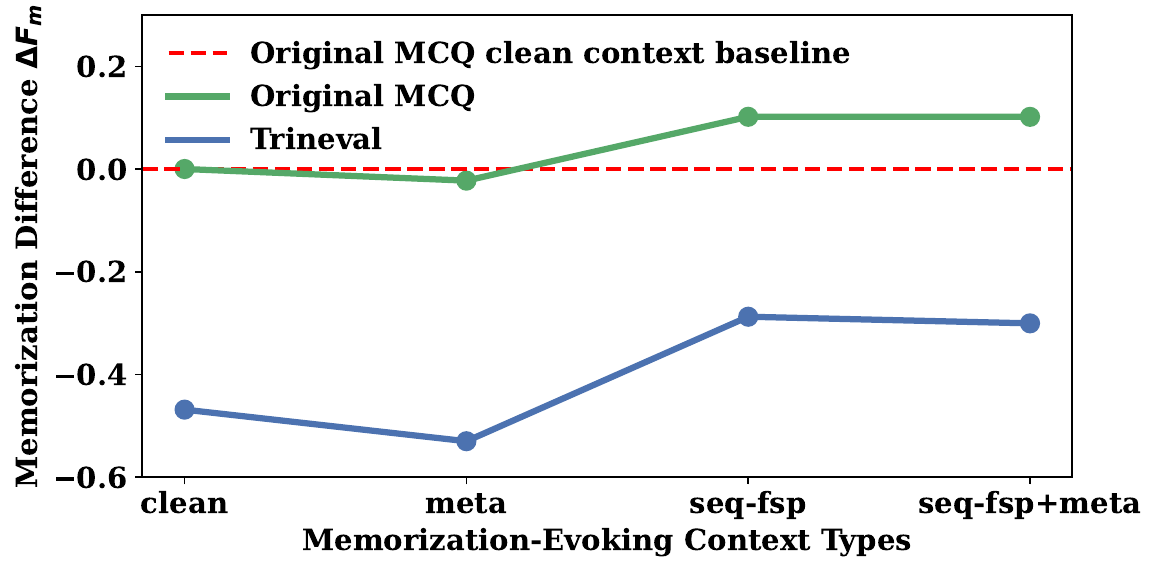} 
        }
    \subfloat[Evoc. with Vicuna and dev-fsp.\label{fig:mempert_subfig3}]{
        \includegraphics[width=0.325\textwidth]{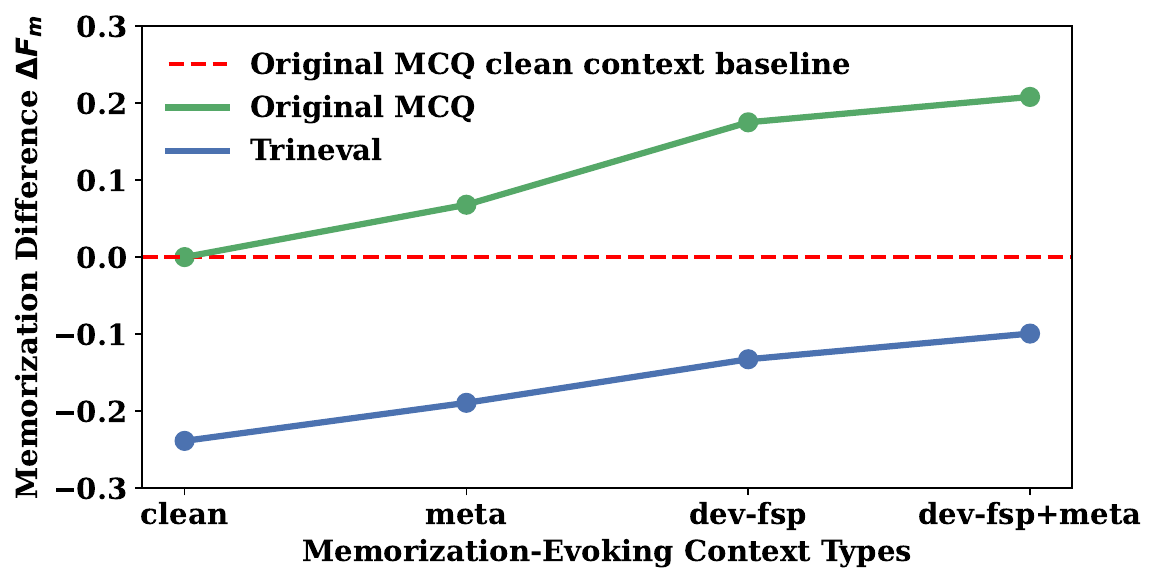} 
        }
    \subfloat[Evoc. with Vicuna and seq-fsp.\label{fig:mempert_subfig6}]{
        \includegraphics[width=0.325\textwidth]{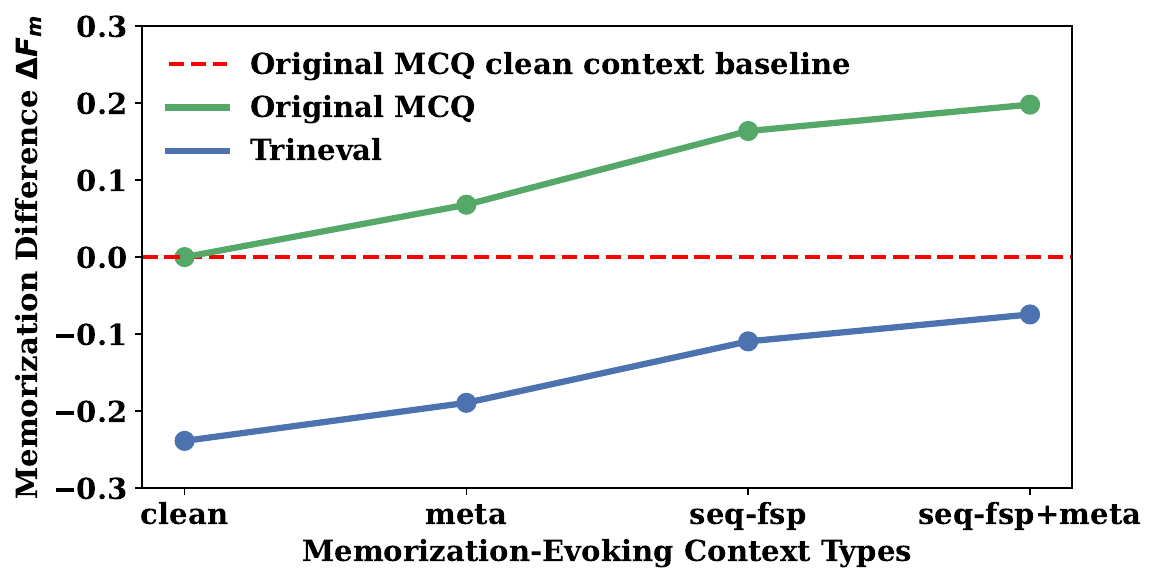} 
        }
    \caption{
    Results of memorization evocation (evoc.) under dataset-related contexts, with green and blue curves referring to the memorization difference $\Delta F_{m}$ in the original and \modelname formats, respectively. In the x-axis, `clean', `meta', `dev-fsp', and `seq-fsp' stand for without dataset-related context, with the name of the dataset, with few-shot prompt from the development set, and with few-shot prompt from the test set ahead of the testing question.
    These results of $\Delta F_{m}$ indicate the growing memorization effect given the increasing dataset-related information in general. However, the $\Delta F_{m}$ by \modelname under the strongest memory evocation context remains consistently lower than the one in the original format, \eg the red dashed line. 
    }\label{fig:validate_robust} 
    \vspace{-2ex}
\end{figure*}


In this subsection, we investigate whether the proposed \modelname reformulation can suppress unnecessary memorization, thereby isolating and revealing the LLMs’ genuine capabilities under various circumstances. Following Bordt et al.~\cite{bordt2024elephants}, we introduce dataset-specific cues into the questioning prompts and assess to what extent \modelname mitigates memorization elicited by such prompts.

We suppose that unintentional data contamination arises from crawling dataset pages (e.g., Hugging Face) during the compilation of LLM pretraining datasets. When researchers are organizing the pretraining corpus, one or more neighboring original data samples would be truncated and concatenated sequentially into a pretraining sample. Thus, according to Carlini's theory~\cite{carlini2022quantifying} and the next-token-prediction pretraining, we believe that offering samples within the same dataset would affect memorization evocation.


Thus, following the previous study~\cite{bordt2024elephants}, to evoke memorization, we embed dataset-related metadata into the input prompts, \ie the dataset name and in-context few-shot examples drawn from the same dataset, to find out if the reformulation of the \modelname framework can eliminate the memorization effect. For this set of experiments, we focus on LlaMA, Mistral, and Vicuna, as there exists verified evidence of data contamination and open-source implementations allow access to token-level output probabilities, enabling us to compute the memorization metric $F_m$. Since $F_m$ lacks a defined absolute zero point indicating complete absence of memorization, we take the $F_m$ value of the original MCQ format without any additional context (\ie, vanilla MCQ) as a reference baseline. We visualize the average change in $F_m$ for each memorization-evoking method relative to this baseline.


Specifically, we hypothesize that unintentional data contamination may occur when LLMs are pretrained on datasets scraped from public sources (\eg, Hugging Face), where nearby examples from the same dataset may be concatenated during corpus construction. Following the pretraining mechanism of next-token prediction and prior findings from Carlini et al.~\cite{carlini2022quantifying}, we posit that the inclusion of dataset-similar samples in the prompt may inadvertently trigger memorization.
Accordingly, we design a sequence of memorization-evoking perturbations by progressively increasing the contextual cues: (1) providing only the dataset name, (2) adding few-shot examples from the same dataset (\ie, samples in the development set or ahead examples in the test set of the same subject of MMLU), and (3) combining both types of context. For each MCQ, we compute the change in $F_m$ relative to the reference baseline, capturing the degree to which memorization is elicited by specific context in original or \modelname formats. Results are illustrated in Fig.~\ref{fig:validate_robust}.


Consistent with the observations by Bordt et al.~\cite{bordt2024elephants}, results show that $F_m$ increases as more dataset-specific context is introduced, indicating stronger memorization effects. However, across all three tested LLMs, the $\Delta F_m$ curve for \modelname remains consistently lower than that of the original MCQ format (\ie the red dashed line) regardless of various context. Notably, even under the strongest memorization-evoking setting, \modelname’s absolute $F_m$ still stays below the vanilla baseline. These findings provide compelling evidence that \modelname significantly mitigates the influence of memorization, effectively disentangling recall from genuine model capability.

\subsection{Q3. \modelname's Findings on LLM Memorization and Genuine Capability Learning}\label{sec:mem_dist}

\begin{figure*}[t]
  \centering
  \subfloat[$F_m$-$F_c$ grouping heatmap via LlaMA on Qwen-extracted triplets.]{%
    \includegraphics[width=0.315\textwidth]{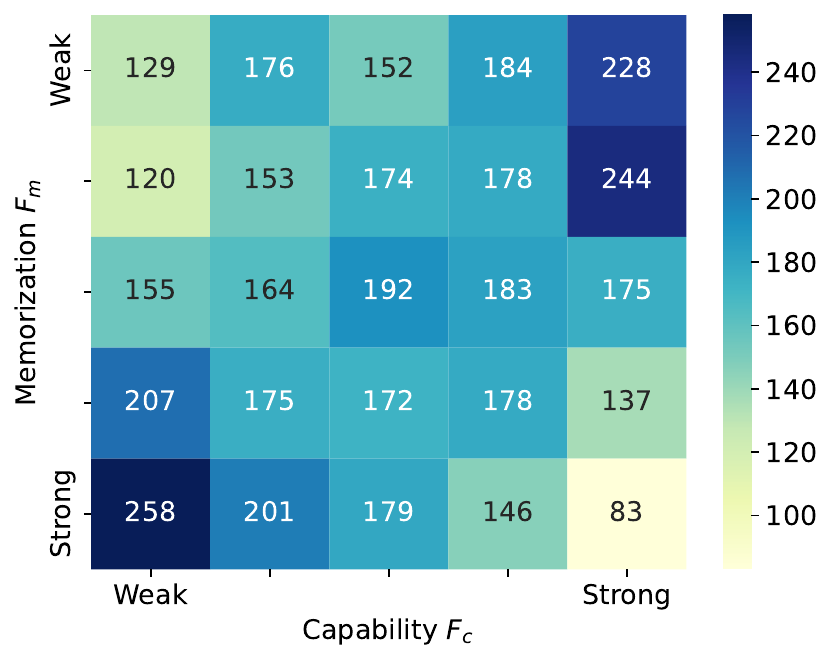}%
  }\hfill
  \subfloat[$F_m$-$F_c$ grouping heatmap via Mistral on Qwen-extracted triplets.]{%
    \includegraphics[width=0.315\textwidth]{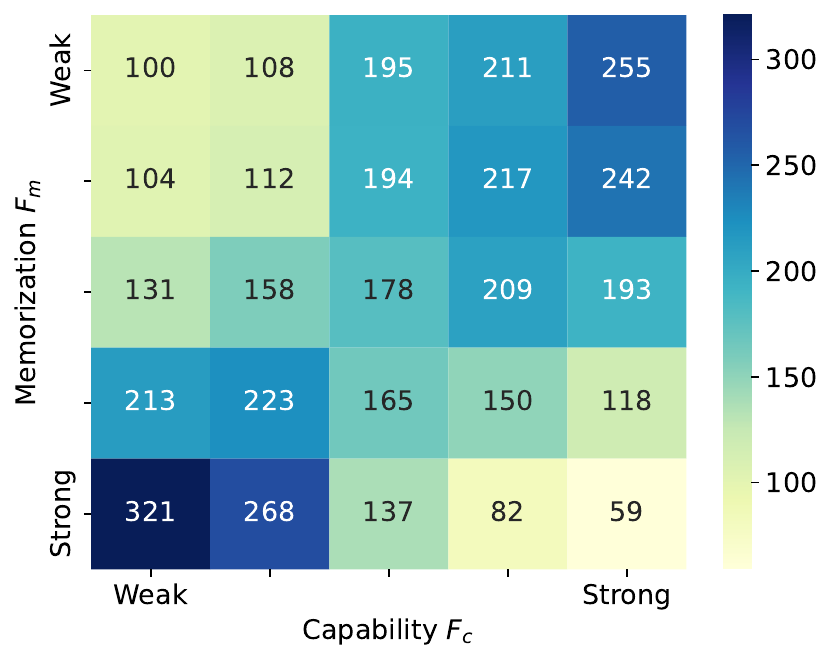}%
  }\hfill
  \subfloat[$F_m$-$F_c$ grouping heatmap via Vicuna on Qwen-extracted triplets.]{%
    \includegraphics[width=0.315\textwidth]{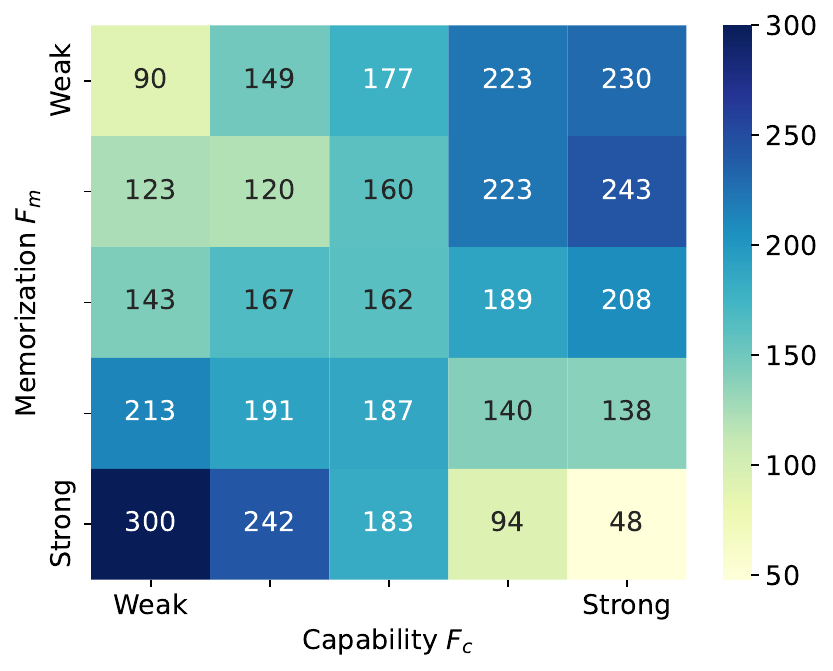}%
  }

  \vfill 

  \subfloat[$F_m$-$F_c$ grouping heatmap via LlaMA on GPT-extracted triplets.]{%
    \includegraphics[width=0.315\textwidth]{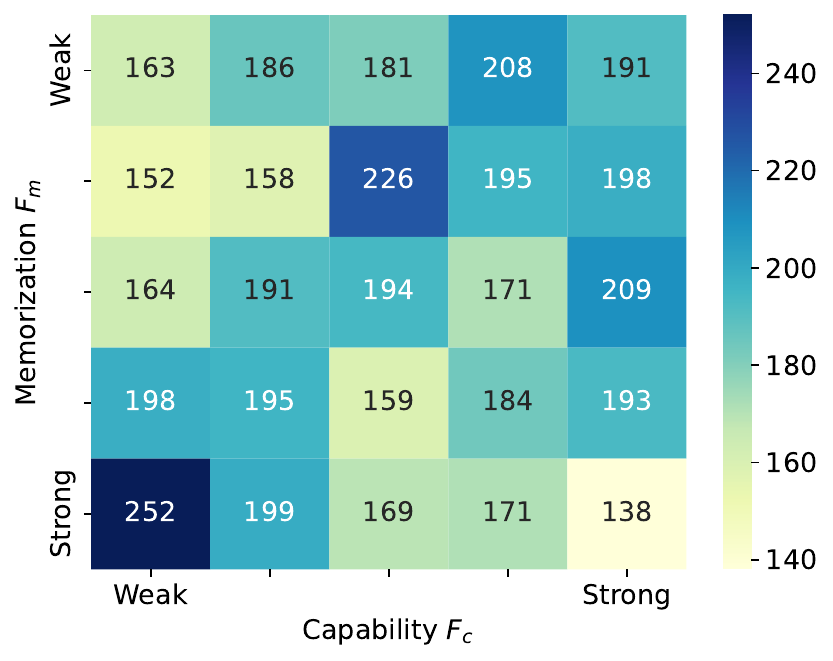}%
  }\hfill
  \subfloat[$F_m$-$F_c$ grouping heatmap via Mistral on GPT-extracted triplets.]{%
    \includegraphics[width=0.315\textwidth]{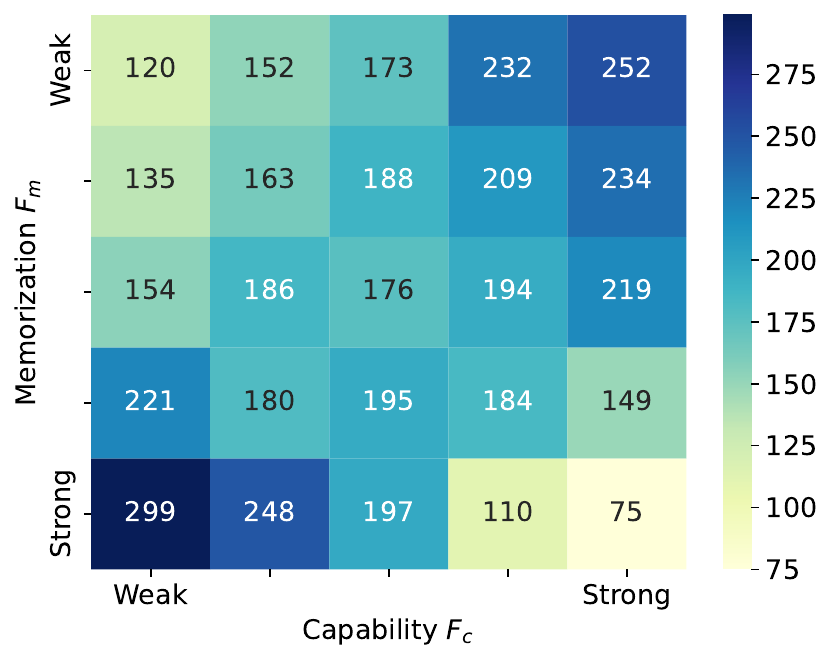}%
  }\hfill
  \subfloat[$F_m$-$F_c$ grouping heatmap via Vicuna on GPT-extracted triplets.]{%
    \includegraphics[width=0.315\textwidth]{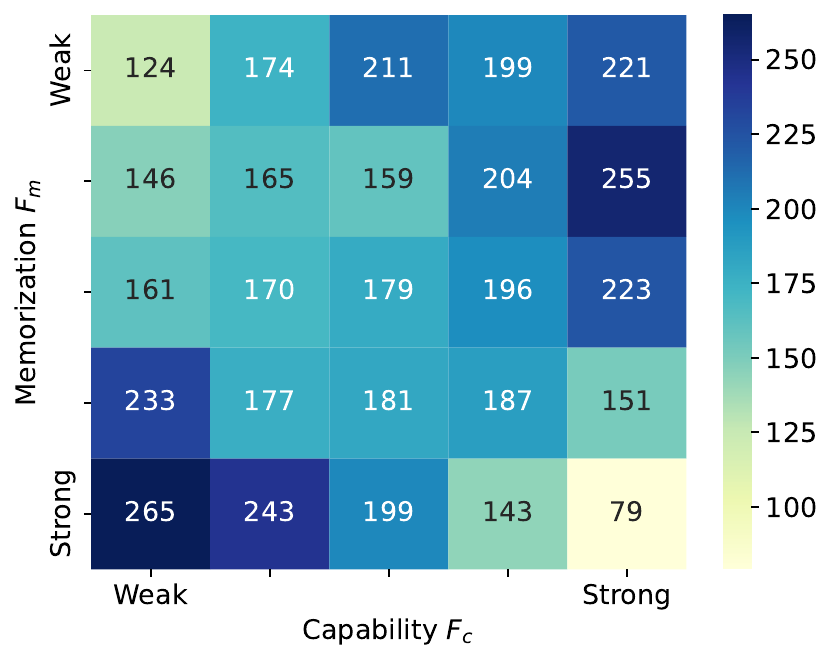}%
  }

  \caption{The distribution of MCQs based on memorization metric $F_m$ vs. capability metric $F_c$. According to the values of $F_m$ and $F_c$, we separate MCQs into 25 groups and visualize the distribution from weak to strong with heatmaps. Same as below.}
  \label{fig:mem_vs_cap}\vspace{-2ex}
\end{figure*}

In this subsection, we aim to examine the relationship between memorization and genuine capability in LLMs, using the metrics $F_m$ (memorization) and $F_c$ (capability). 

\subsubsection*{Q3.1 \modelname's findings on MCQs}
Since commercial proprietary models do not expose full vocabulary-level output probabilities, we focus on open-source LLMs for computing these metrics. After computing the $F_m$ and $F_c$ scores for all qualified MCQs in \modelname format, we divide the values of $F_m$ and $F_c$ into five equal intervals, respectively, resulting in 25 distinct groups of MCQ samples based on their joint distribution. We then use heatmaps to visualize the distribution of samples across these groups, revealing the relationship between memorization and capability for each tested model.

As shown in Fig.\ref{fig:mem_vs_cap}, the majority of MCQs cluster in the lower-left and upper-right corners of the heatmap. For instance, in the results of LLaMA using Qwen-extracted triplets, 38.57\% of MCQs fall within the combined lower-left and upper-right $2 \times 2$ grid regions, yielding a Pearson correlation of -0.7755 (p-value $< 0.05$) for $F_m$ vs. $F_c$ of examples within the two square regions; expanding to the $3 \times 3$ regions increases coverage to 74.17\%, with a stronger correlation of -0.8124 (p-value $< 0.05$). Similarly, for Mistral on Qwen triplets, 44.90\% of MCQs lie in the two $2 \times 2$ corner regions with a correlation of -0.8722 (p-value $< 0.05$), and 80.82\% in the two $3 \times 3$ corner regions with a correlation of -0.8794 (p-value $< 0.05$). 

\begin{figure*}[t]
    \centering
    \subfloat[Grouping heatmap via LlaMA3.1-8B on Qwen3-extracted triplets.]{%
        \includegraphics[width=0.315\textwidth]{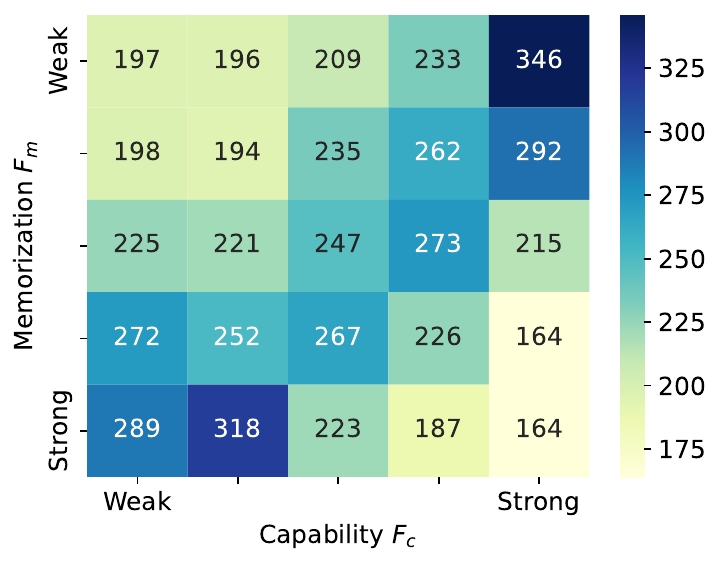}%
    }\hfill
    \subfloat[Grouping heatmap via Qwen3-14b on Qwen3-extracted triplets.]{%
        \includegraphics[width=0.315\textwidth]{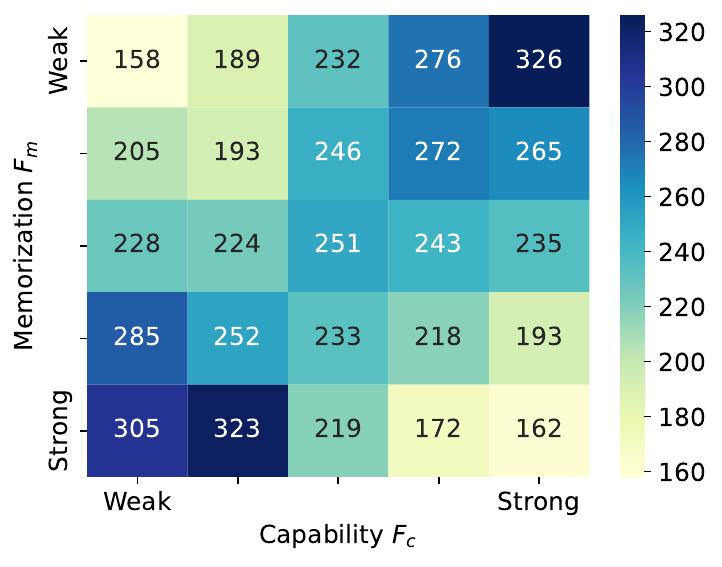}%
    }\hfill
    \subfloat[Grouping heatmap via Qwen3-32b on Qwen3-extracted triplets.]{%
        \includegraphics[width=0.315\textwidth]{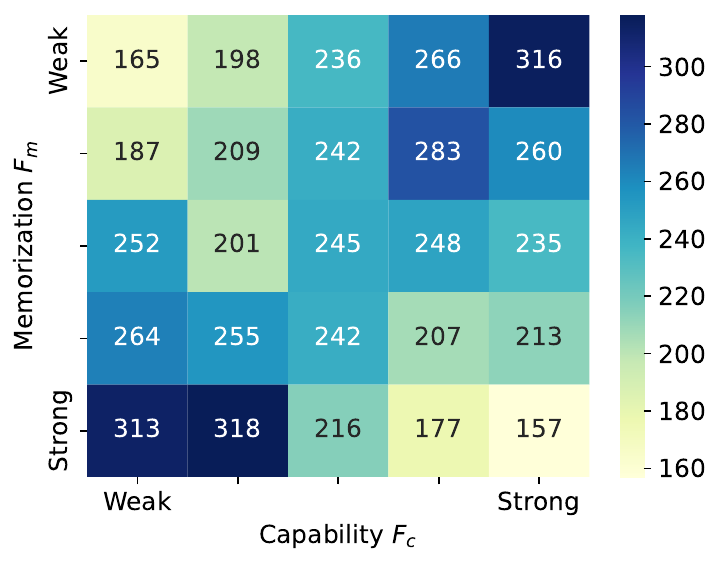}%
    }

    \vfill 

    \subfloat[Grouping heatmap via LlaMA3.1-8B on Dpsk-extracted triplets.]{%
        \includegraphics[width=0.315\textwidth]{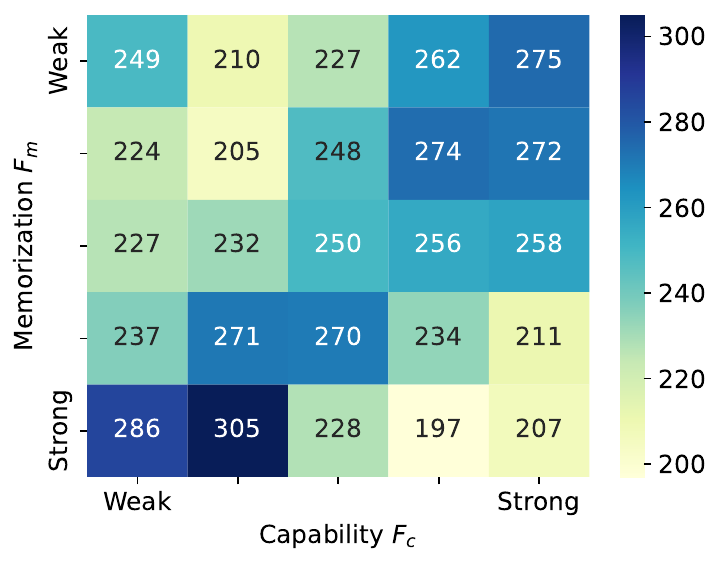}%
    }\hfill
    \subfloat[Grouping heatmap via Qwen3-14b on Dpsk-extracted triplets.]{%
        \includegraphics[width=0.315\textwidth]{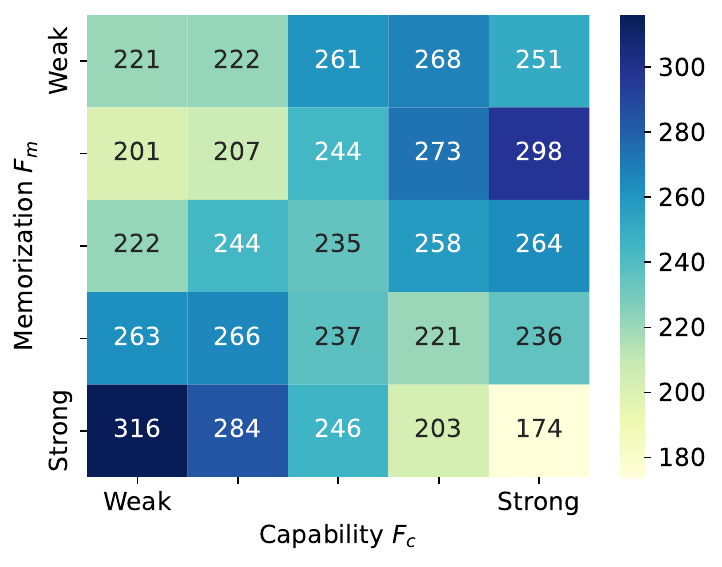}%
    }\hfill
    \subfloat[Grouping heatmap via Qwen3-32b on Dpsk-extracted triplets.]{%
        \includegraphics[width=0.315\textwidth]{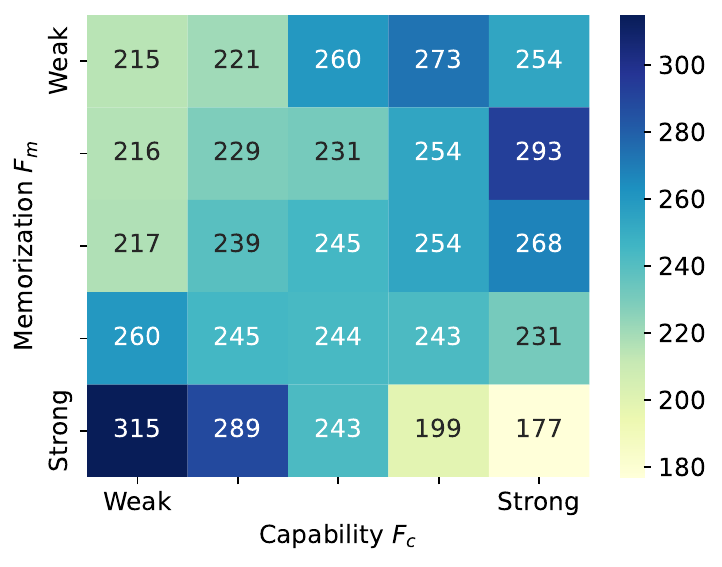}%
    }

    \caption{The distribution of MCQs based on memorization metric $F_m$ vs. capability metric $F_c$ with latest larger models.}\vspace{-2.5ex}
    \label{fig:mem_vs_cap_1}
\end{figure*}

We analyze the distribution in detail by revealing the ratio of MCQs within the upper right and lower left 2 $\times$ 2 and 3 $\times$ 3 squares as well as the Pearson correlations between the $F_m$ and $F_c$ of these MCQs. We take the MCQs within the lower-left 2 $\times$ 2 squares in Fig.~\ref{fig:mem_vs_cap} as the ones memorized by rote and compute the ratio of 20.5\% by averaging the percentage of all evaluated LLMs. Similarly, 19.6\% is computed as the ratio of MCQs within the upper-right 2 $\times$ 2 squares with genuine capability. All of these MCQs also exhibit a tendency towards a negative correlation between the capabilities and memorization of LLMs (all Pearson correlation values are computed with p-values at the 0.05 level).
Additional statistics results are reported in Tab. III in Appendix C. 
Further, we also plot the curve of the separation of the rote memorization and genuine capability learning MCQs in Appendix C. With each progressing $F_m$ as threshhold, the tested MCQs exhibit prominent $F_c$ gaps between the two groups, showing significant difference in learning extremes.

These findings consistently indicate a negative correlation between memorization and capability: \textbf{MCQs with lower memorization scores are more likely to reflect genuine problem-solving capability, whereas high memorization levels correspond to degraded performance}, suggesting overfitting or shallow recall. Still, we would argue that knowledge acquisition in LLMs is likely a continuous spectrum rather than a strict binary. Our preliminary investigation also highlighted the limitations of a discrete classification, and thus, we apply the continuous quantification metric $F_m$ and $F_c$. As seen in the heatmaps, some questions fall in intermediate regions, indicating varying degrees of memorization and understanding. Thus, while we focus on the two extremes for analytical clarity, we agree that other intermediate or hybrid categories may exist.

\subsubsection*{Q3.2 Can \modelname's findings be extended to latest larger LLMs?}

Based on previous research~\cite{touvron2023llama, sainz2023nlp}, the aforementioned experiments were conducted in a relatively deterministic experimental environment, \ie confirmed data contamination evidence. Therefore, we attempt to extend the above results to the latest larger LLMs. Here, we mainly reformulate the MMLU MCQs based on Dpsk and Qwen3, obtaining 6,115 and 5,905 MCQs, respectively, and verify the results based on LlaMA3.1-8B, Qwen3-14B, and Qwen3-32B. The experimental procedure remains identical to that described in the preceding experiment. Results are shown in Fig.~\ref{fig:mem_vs_cap_1}. We can see that the majority of MCQs also cluster in the lower-left and upper-right corners of the heatmap, which is evident and consistent across our tests. For instance, in the results of LlaMA3.1 using Qwen3-extracted triplets, 38.34\% of the MCQs fall within the two opposite corner regions of the $2 \times 2$ grid. For the examples in these two regions, the Pearson correlation between $F_m$ and $F_c$ is -0.7771 (p-value $< 0.05$). Extending the analysis to the $3 \times 3$ corner regions increases coverage to 78.34\%, yielding a correlation of 0.5506 (p-value $< 0.05$). Similarly, for Qwen-14B on Qwen triplets, 39.04\% of the MCQs lie in the two $2 \times 2$ corner regions, with a correlation of -0.7260 (p-value $< 0.05$); and 79.02\% fall in the two $3 \times 3$ corner regions, with a correlation of -0.4905 (p-value $< 0.05$).

\begin{figure*}[ht]
    \centering
    \subfloat[Grouping heatmap via LlaMA on Qwen3-extracted triplets.]{%
        \includegraphics[width=0.315\textwidth]{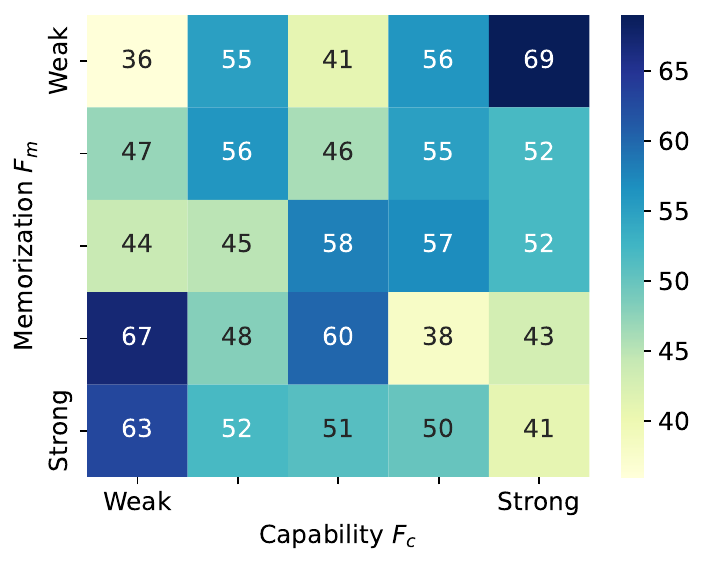}%
    }\hfill
    \subfloat[Grouping heatmap via Mistral on Qwen3-extracted triplets.]{%
        \includegraphics[width=0.315\textwidth]{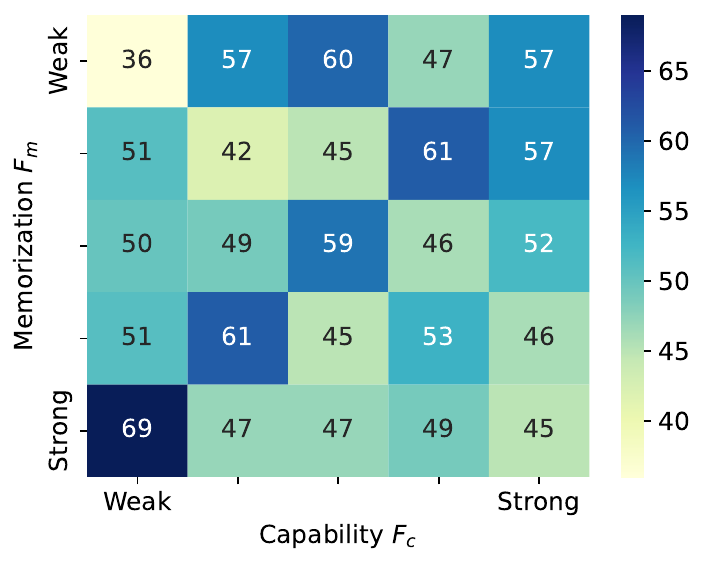}%
    }\hfill
    \subfloat[Grouping heatmap via Vicuna on Qwen3-extracted triplets.]{%
        \includegraphics[width=0.315\textwidth]{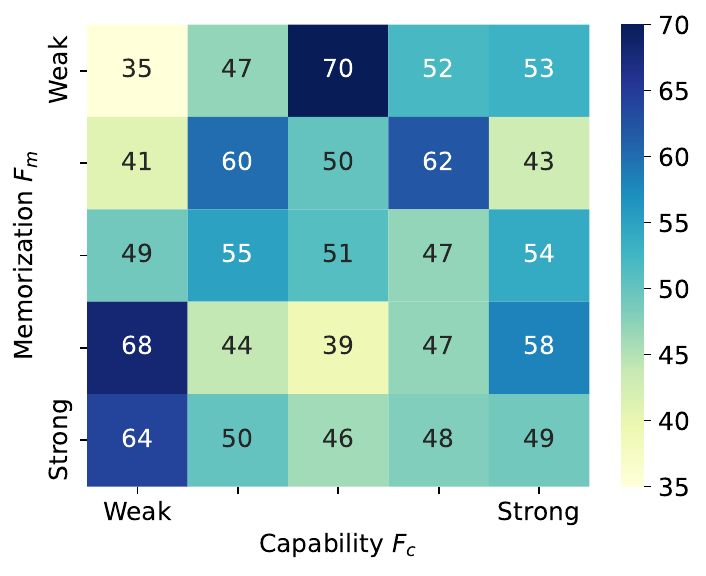}%
    }

    \vfill 

    \subfloat[Grouping heatmap via LlaMA on Dpsk-extracted triplets.]{%
        \includegraphics[width=0.315\textwidth]{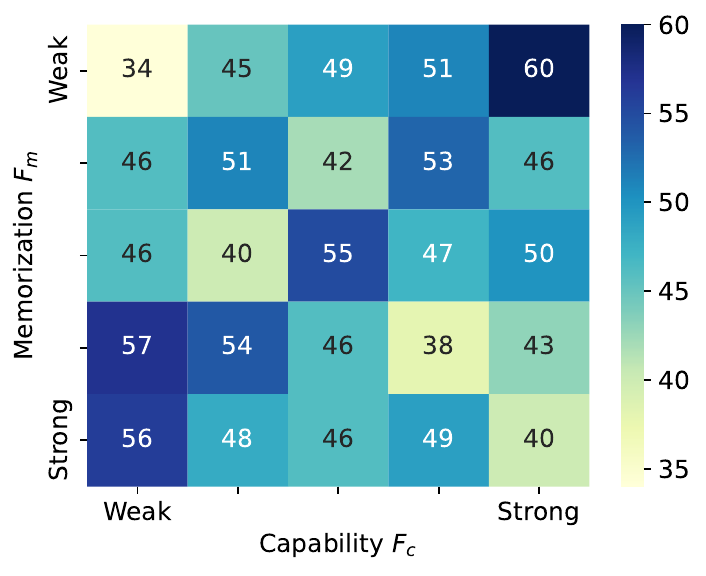}%
    }\hfill
    \subfloat[Grouping heatmap via Mistral on Dpsk-extracted triplets.]{%
        \includegraphics[width=0.315\textwidth]{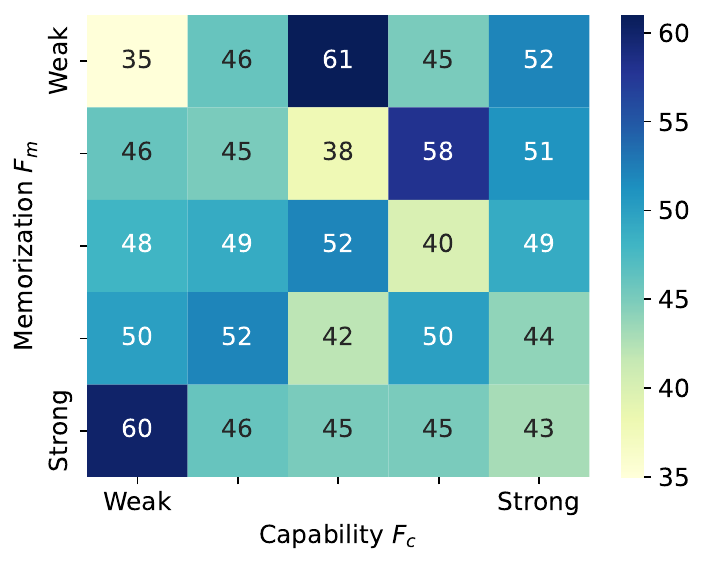}%
    }\hfill
    \subfloat[Grouping heatmap via Vicuna on Dpsk-extracted triplets.]{%
        \includegraphics[width=0.315\textwidth]{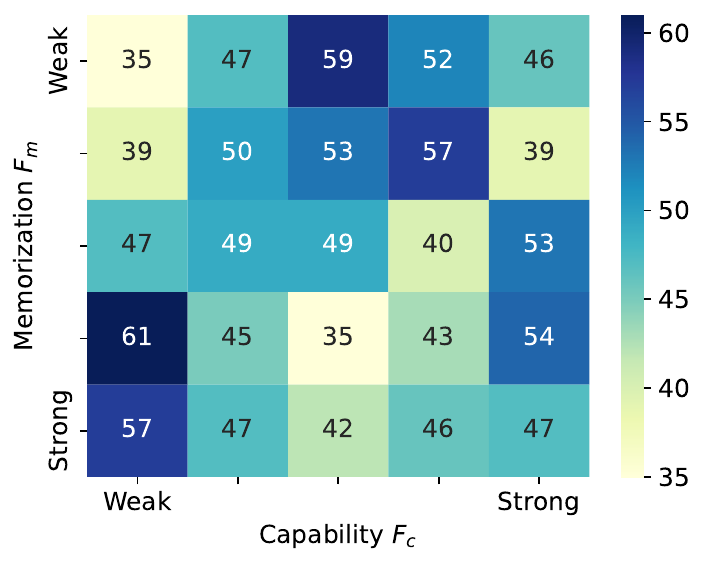}%
    }

    \caption{The distribution of OEQs based on memorization metric $F_m$ vs. capability metric $F_c$.}
    \label{fig:mem_vs_cap_oeq}\vspace{-2ex}
\end{figure*}

\begin{figure*}[htbp]
  \centering
  \subfloat[Grouping heatmap via LlaMA3.1-8B on Qwen3-extracted triplets.\label{fig:subfig1}]{%
    \includegraphics[width=0.315\textwidth]{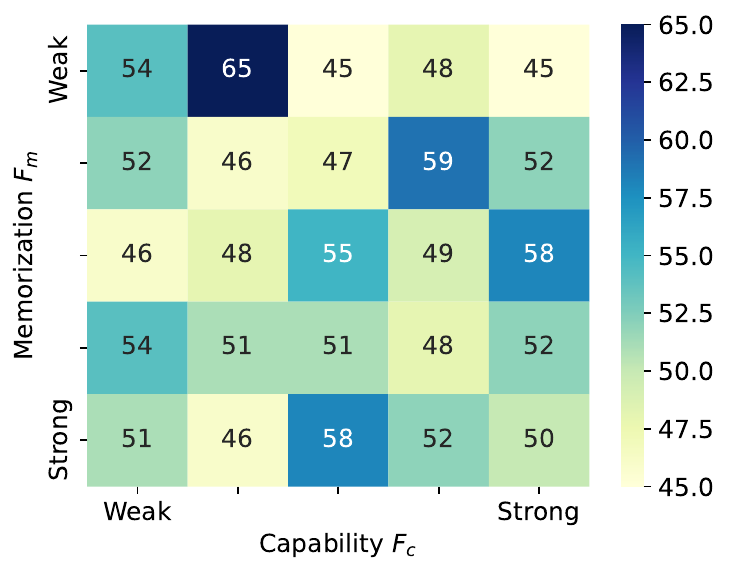}%
  }\hfill
  \subfloat[Grouping heatmap via Qwen3-14b on Qwen3-extracted triplets.\label{fig:subfig2}]{%
    \includegraphics[width=0.315\textwidth]{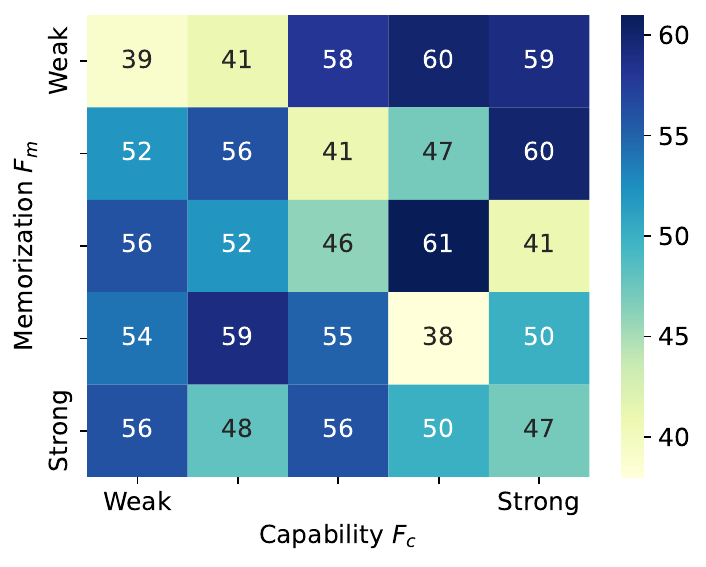}%
  }\hfill
  \subfloat[Grouping heatmap via Qwen3-32b on Qwen3-extracted triplets.\label{fig:subfig3}]{%
    \includegraphics[width=0.315\textwidth]{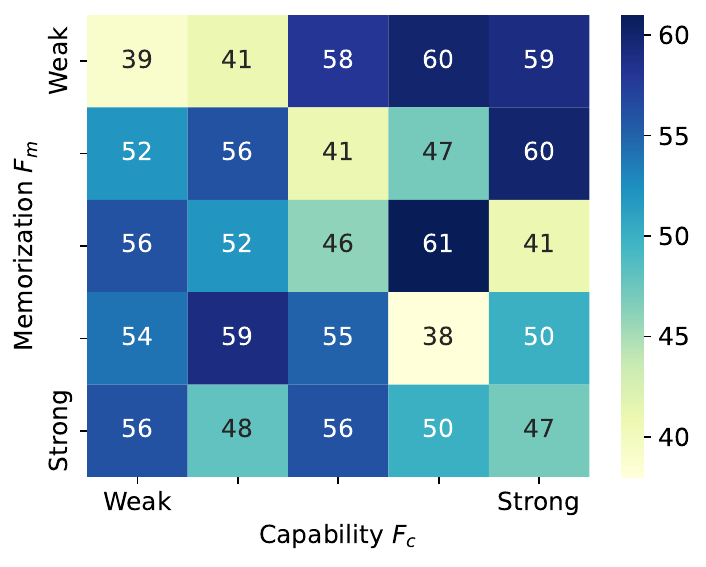}%
  }

  \vfill  

  \subfloat[Grouping heatmap via LlaMA3.1-8B on Dpsk-extracted triplets.\label{fig:subfig4}]{%
    \includegraphics[width=0.315\textwidth]{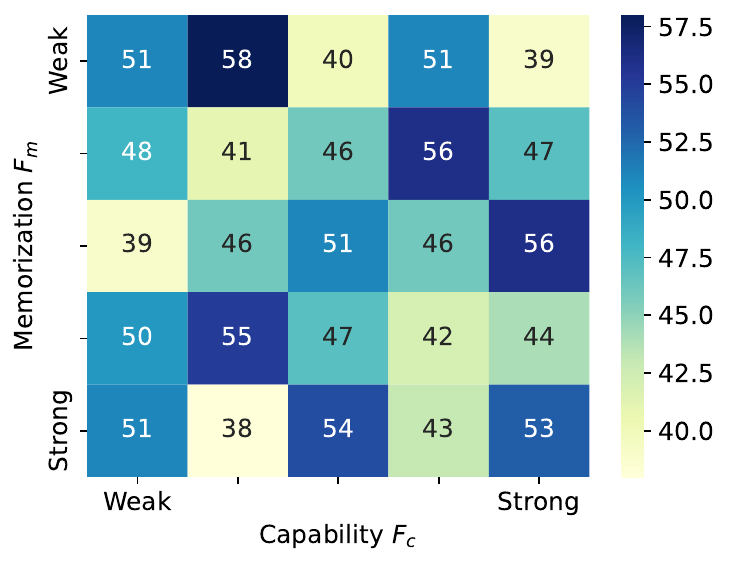}%
  }\hfill
  \subfloat[Grouping heatmap via Qwen3-14b on Dpsk-extracted triplets.\label{fig:subfig5}]{%
    \includegraphics[width=0.315\textwidth]{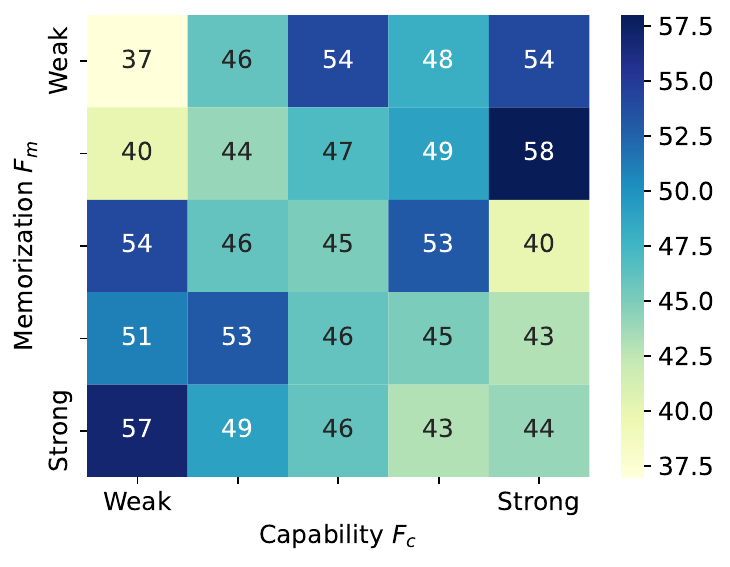}%
  }\hfill
  \subfloat[Grouping heatmap via Qwen3-32b on Dpsk-extracted triplets.\label{fig:subfig6}]{%
    \includegraphics[width=0.315\textwidth]{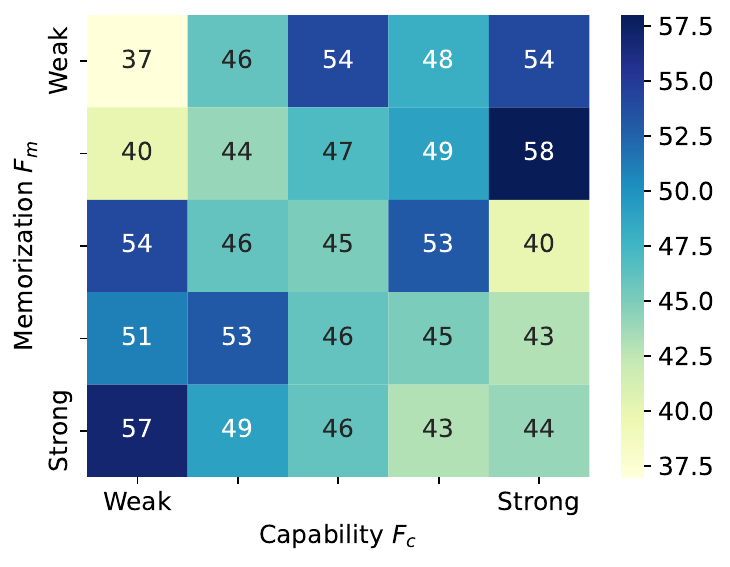}%
  }

  \caption{The distribution of OEQs based on memorization metric $F_m$ vs. capability metric $F_c$ with latest larger models.}\vspace{-2ex}
  \label{fig:mem_vs_cap_oeq_1}
\end{figure*}

\subsubsection*{Q3.3 Can \modelname's findings be extended to OEQs?}

Subsequently, we pose a new question: can the above findings on MCQs be extended to OEQs? To answer this question, we apply \modelname on the GSM8K dataset. For $F_m$, instead of the original option content in MCQs, we select the bottom K\% token probabilities of the official CoT seuqence for each OEQ to evluate the memorization of LLMs. As to $F_c$, given that GSM8K is a mathematics problem benchmark with a definite answer, we utilize the perplexity of the answer sequence. For open-ended questions (OEQs) that admit multiple correct answer formulations and require semantic correctness evaluation, we believe that our \modelname can also be readily adapted to corresponding measurement. We utilize TrinEval to reformulate the OEQs and obtain 1282 qualified questions and use them to evaluate LlaMA, Mistral, Vicuna, as well as latest larger models, Llama3.1-8B, Qwen3-14B, and Qwen3-32B. The results are shown in Fig.~\ref{fig:mem_vs_cap_oeq} and Fig.~\ref{fig:mem_vs_cap_oeq_1}.

We surprisingly find that on OEQs, most LLMs still exhibit the same phenomena, \ie OEQs with lower memorization scores are more likely to reflect genuine problem-solving capability, whereas high memorization levels correspond to degraded performance. Among them, for LLaMA on the triplet questions extracted by Dpsk, 18.04\% of the questions exhibited rote memorization, while 17.62\% were genuinely mastered. For Qwen3‑32B on the same Dpsk‑extracted triplets, 17.62\% of the questions showed rote memorization, and 17.53\% were genuinely mastered. However, for some results, for instance, LLaMA3.1‑8B on the triplet extracted by Qwen3, only 15.76\% of the questions were classified as rote memorization, falling below the 16\% baseline in the $5 \times 5$ heatmap. We attribute this to the following reasons. First, comparing Fig.~\ref{fig:mem_vs_cap_oeq} and Fig.~\ref{fig:mem_vs_cap_oeq_1}, we observe that the distinction between rote memorization and genuine capability learning is less pronounced in the latest larger LLMs than in other LLMs. This may be because, as suggested by recent research~\cite{hao2025reformulation, fujii2025rewriting, ovadia2023fine}, LLMs are increasingly designed to be trained on reformulated corpus during pretraining, thereby reducing the formation of such rote memorization. Second, Several studies~\cite{liu2026format} have also found that overly formatted corpora can lead to poorer generalization in LLMs. Thus, compared to MCQs, OEQs are less structured and thus may be less susceptible to rote memorization. Nevertheless, the conclusions drawn from most LLMs remain consistent with our earlier experimental findings on MCQs, suggesting that LLMs also exhibit rote memorization in the context of OEQs.

\subsubsection*{Q3.4 How to interpret \modelname's findings?}

To interpret these results, we draw an analogy between LLM behavior and the human memory system, which comprises two key components: Short-Term Memory (STM) and Long-Term Memory (LTM)~\cite{shiffrin2003chapter}.
Neurobiological studies reveal that STM relies on transient synaptic protein synthesis with limited temporal persistence and functional scalability. In contrast, LTM is constructed through stabilized neuronal memory traces that constitute an enduring knowledge framework. This neural architecture not only supports STM operations as a cognitive substrate but also enables sophisticated information generalization across diverse contexts. 
This dichotomy aligns with recent observations in LLMs. As shown by Allen et al.~\cite{allen2023physics} and Ovadia et al.~\cite{ovadia2023fine}, models trained on corpora with diverse rephrasings exhibit stronger generalization than those trained solely on original formulations. When trained on a single fixed corpus format, LLMs, due to their strong memorization capacity, tend to activate their STM system to memorize at the token level, capturing surface patterns rather than abstract knowledge. In this sense, \textbf{LLMs are potential rote learners}.
Reformulations like \modelname appear to facilitate the encoding of knowledge in a more principled, structured and reusable form, akin to LTM, thereby supporting more generalized and robust evaluation.

\begin{figure}[!ht]
  \includegraphics[width=\linewidth]{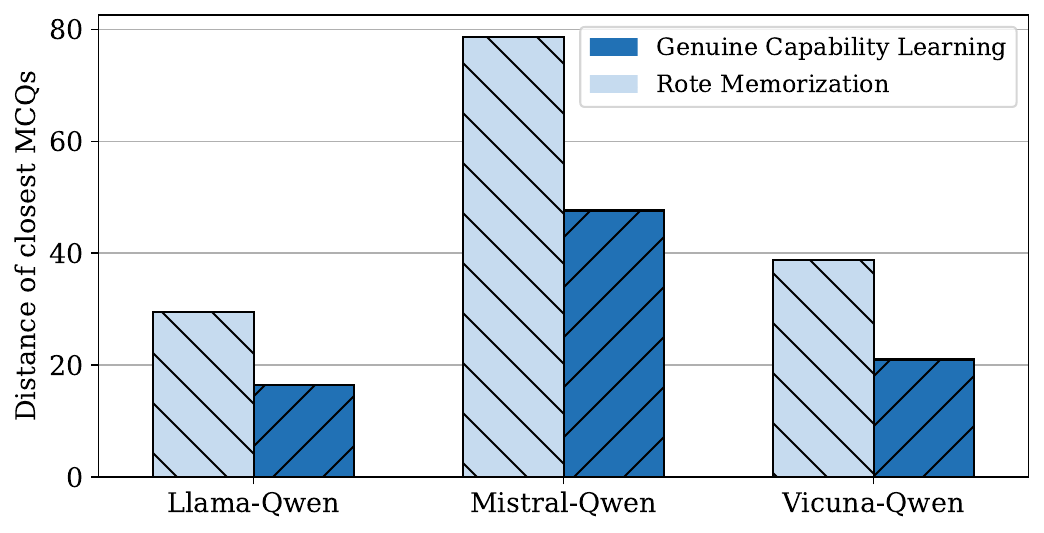} 
  \includegraphics[width=\linewidth]{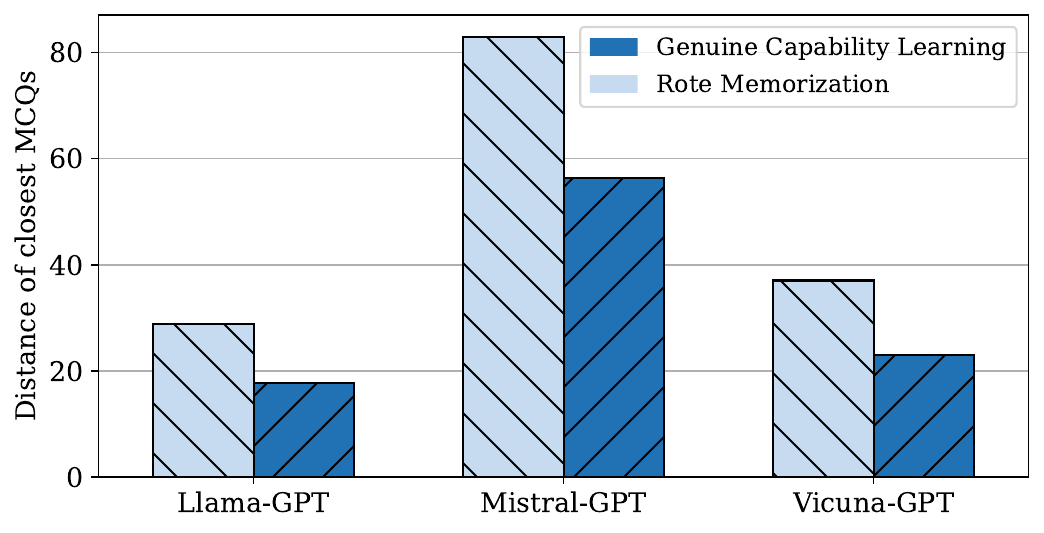}
  \caption {Averaged distance of each MCQs between the closest $1\%$ MCQs' embeddings. `Rote Memorization' refers to MCQ within lower left $2 \times 2$ squares that typically exhibits a high memorization metric $F_m$ and a low capability $F_c$ while `Genuine Capability Learning' stands for MCQ lies within the upper right $2 \times 2$ squares 
  with low $F_m$ but high $F_c$.}
  \label{fig:dist}\vspace{-2ex}
\end{figure}

To further validate our hypothesis, we analyze the semantic proximity of MCQs within the qualified MMLU dataset by computing their embeddings and measuring the average distance to their closest 1\% samples. 
Our underlying assumption is that LLMs could better master a knowledge point if it is expressed in various formats in the training corpus. As a result, these knowledge `rephrases' are also encoded densely via LLMs.
As there are 57 different subjects within the MMLU dataset, we argue that unrelated sequences would lead to increased embedding distance even though they are among the mastered knowledge points. Here, we try to filter out the unrelated samples for each MCQ and thus keep the closest samples at the 1e-2 level in order to make sure there are not too many unrelated samples incorporated. To highlight this result more prominently, we employ a relatively stringent filtering strategy and made it 1\% of the closest samples, aiming to retain the samples that are semantically correlated in knowledge description.
We then compare the mean embedding distances of MCQs located in the lower-left (strong memorization, weak capability) and upper-right (weak memorization, strong capability) $2 \times 2$ regions in Fig.~\ref{fig:mem_vs_cap}. The corresponding results are presented in Fig.~\ref{fig:dist}, where RM stands for rote memorization, and GCL stands for genuine capability learning. In order to provide a more robust result, we also present results obtained under more lenient data filtering criteria, such as thresholds of 3\% and 5\% in Tab. II in Appendix D. Interestingly, we observe that the average embedding distance among the Genuine Capability Learning MCQs is much lower than (nearly half) that of the Rote Memorization MCQs. This suggests that memorized MCQs are more sparsely distributed in the embedding space, while non-memorized, capability-driven MCQs tend to form tighter semantic clusters. The observation also aligns well with the cognitive analogy of STM and LTM: rote memorization leads to fragmented, context-specific encoding, whereas genuine capability emerges from structured, reusable representations. Also, we can see that the more samples we incorporated, the higher the average distance of the closest embeddings grows. Still, though we increase the filtering threshold and the distance gap between the RM and the GCL is narrowing, we can still find that the distance between the closest rote memorization MCQ embeddings is more than the distance between the closest genuine capability learning MCQ embeddings. As a conclusion, though it is widely believed that memorization may lead to better but cheating performance of LLMs, we find that the more LLMs memorize, the worse they are at solving problems.

\section{Conclusion}


This study provided a novel perspective on benchmark contamination in LLM evaluation, reframing it as an inherent aspect of learning. This perspective led us to explore the relationship between memorization and genuine capability in LLMs.
Through our empirical investigation, we observed a surprising result: LLMs performed worse on memorized benchmarks compared to those not, suggesting that superficial memorization may undermine problem-solving ability rather than enhance it. 
This finding also implies the existence of two distinct learning paradigms in LLMs: rote memorization and genuine capability learning.
To disentangle them, we proposed \modelname, an evaluation method reformulating MCQs and OEQs into a knowledge-centric trinity and separating the influence of memorization from genuine knowledge application. 
Experiments validated both knowledge-preserving and memorization-reducing properties of \modelname. Based on that, \modelname reveals the in-robustness of LLMs’ knowledge learning, \eg popular open-source LLMs memorize 20.5\% of knowledge points by rote 
in MMLU. We also discussed the generalizability of our approach to other forms of questions (\eg open-ended questions) in Sec~\ref{sec:exp_mem}.
As such, we believe this work lays the groundwork for future studies on improving LLM knowledge robustness and a more thorough evaluation.

\appendices
{\section{Details of the extracting prompts and the extracted (memorized) MCQs}
\label{apd:pre_exp}

In this section, we elaborate on the details of the processed dataset and the prompts for extraction. MCQs from some subjects contain similar or identical options\footnote{\eg, the options of MCQs in the subject, moral\_scenarios, are all identical (`Wrong, Wrong', `Wrong, Not wrong', `Not wrong, Wrong' and `Not wrong, Not wrong').}. With the provided 5-shot prompt, options of MCQs from these subjects can be easily extracted, leading to a high False-positive ratio. In order to avoid the influence of the few-shot prompt on the option extraction, we eliminate MCQs in which any of the options have appeared twice in the dataset. After deduplication, we obtain 14,006 MCQs for evaluation. The detailed prompt and statistics are shown in the following Fig.~\ref{fig:pre_exp_prompt} and Tab.~\ref{tab:pre_exp_sta}.

\begin{table}[h]
\centering
\begin{tblr}{
  cells = {c},
  cell{2}{1} = {r=3}{},
  cell{5}{1} = {r=3}{},
  cell{8}{1} = {r=3}{},
  vline{2-3} = {1-2,5,8}{},
  vline{2-3} = {3-4,6-7,9-10}{},
  hline{1-2,5,8,11} = {-}{},
}
\textbf{Model} & \textbf{Subset} & \textbf{Simple} & \textbf{Pro} & \textbf{MMLU} \\
{Llama}   & {memorized}   & 912             & 70               & 982          \\
                 & {non-mem.} & 6,548            & 6,476             & 13,024        \\ 
                 & {all}         & 7,460            & 6,546             & 14,006        \\
{Mistral} & {memorized}   & 879             & 36               & 915          \\
                 & {non-mem.} & 6,581            & 6,510             & 13,091        \\ 
                & {all}         & 7,460            & 6,546             & 14,006        \\ 
{Vicuna}  & {memorized}   & 893             & 16               & 909          \\
                 & {non-mem.} & 6,567            & 6,530             & 13,097        \\
                 & {all}         & 7,460            & 6,546             & 14,006        
\end{tblr}
\caption{Statistics of memorized and non-memorized questions by Llama2-7B, Mistral-7B-v0.2, and Vicuna-v1.5-7B in MMLU.}\label{tab:pre_exp_sta}
\end{table}

\section{Details of \modelname}\label{apd:reorg}

In this section, we introduce the details of the reflection part of the proposed \modelname. The extraction and reflection prompts used are shown in Fig.~\ref{fig:reorg}, Fig.~\ref{fig:reorg_verdict}, and Fig.~\ref{fig:reorg_reflect} below. The reformulated example is shown in Fig.~\ref{fig:mcq_form}.

\section{Detailed results of memorization v.s. capability}
\label{apd:mem_cap}

In this section, we exhibit the detailed results of the Q3. What does \modelname reveal about the memorization v.s. the capability of LLMs, as shown in Tab.~\ref{tab:mem_cap}.

Further, inspired by the Precision-Recall Curve, we take each unique $F_m$ of the qualified MCQs as the threshold to separate them as the Memorized and Capable MCQs. For each separation, we compute the probability of whether the $F_c$ of a randomly selected Capable MCQ exceeds the $F_c$ of a randomly selected Memorized MCQ and plot them as the blue curve. We also compute the T-test p-value between the $F_c$s of the Memorized MCQs and Capable MCQs as the green curve. The results are shown in Fig.~\ref{fig:conti_mem_cap}. For the second row, we filter out the MCQs within the upper left and lower right $2\times 2$ squares. From the figure, we observe that over a relatively long segment in the middle of the x-axis threshold range, the probability remains at a comparatively high value, while the p-value stays below 0.05. From this, we can conclude that $F_m$ can distinguish between MCQs with high $F_c$ and those with low $F_c$ with a negative correlation at a high confidence level. 

\section{Detailed results of Q3.4}

To highlight this result more prominently, we employed a relatively stringent data filtering strategy in the paper and made it 1\% of the closest samples in Sec.~IV-D. In this section, in order to provide a more robust result, we also present results obtained under more lenient data filtering criteria, such as thresholds of 3\% and 5\%. The results are shown in Tab.~\ref{tab:mem_dist}, where RM stands for rote memorization, and GCL stands for genuine capability learning.


\begin{table}[h!]
\centering\scriptsize
\begin{tblr}{
  cells = {c},
  cell{1}{2} = {c=2}{},
  cell{1}{4} = {c=2}{},
  cell{1}{6} = {c=2}{},
  vline{2-3,5} = {1}{},
  vline{2,4,6} = {2-8}{},
  hline{1-3,9} = {-}{},
}
\textbf{Threshold}    & \textbf{1\% } &              & \textbf{3\% } &              & \textbf{5\% } &              \\
\textbf{Subset}       & \textbf{RM}   & \textbf{GCL} & \textbf{RM}   & \textbf{GCL} & \textbf{RM}   & \textbf{GCL} \\
\textbf{Llama-Qwen}   & 29.538        & 16.501       & 32.718        & 17.281       & 33.790        & 18.060       \\
\textbf{Llama-GPT}    & 28.925        & 17.777       & 31.327        & 18.275       & 32.315        & 19.068       \\
\textbf{Mistral-Qwen} & 78.663        & 47.648       & 86.003        & 52.995       & 89.598        & 55.525       \\
\textbf{Mistral-GPT}  & 82.932        & 56.375       & 90.597        & 62.112       & 94.384        & 64.820       \\
\textbf{Vicuna-Qwen}  & 38.761        & 21.006       & 41.396        & 23.189       & 42.490        & 24.185       \\
\textbf{Vicuna-GPT}   & 37.037        & 22.996       & 41.396        & 23.189       & 42.490        & 24.185       
\end{tblr}
\caption{Result of averaged embedding distance of the closest MCQs under different thresholds}\label{tab:mem_dist}
\end{table}

We can see that, as we concluded in the main part, the more samples we incorporated, the higher the average distance of the closest embeddings grows. Still, though we increase the filtering threshold and the distance gap between the RM and the GCL is narrowing, we can still find that the distance between the closest rote memorization MCQ embeddings is more than the distance between the closest genuine capability learning MCQ embeddings. This proves that the reported results are robust and solid.

\begin{figure*}
\centering
\begin{preouterbox}
    \noindent
    \begin{minipage}[t]{0.48\textwidth}
        \begin{preextractbox}
\textit\scriptsize{You are an expert of multiple choice questions of the MMLU dataset. The following are multiple-choice questions (with answers) about [subject].}

~\

\noindent\textit{[examples]}

~\

\noindent\textit{[question]}

\noindent\textit{Options:}

\noindent\textit{A. }

~

~

~

~

~

\end{preextractbox}
\end{minipage}
    \hfill
    \begin{minipage}[t]{0.48\textwidth}
        \begin{preperformancebox}
\textit\scriptsize{You are an expert of multiple choice questions of the MMLU dataset. The following are multiple-choice questions (with answers) about [subject].}

~\

\noindent\textit{[examples]}

~\

\noindent\textit{[question]}

\noindent\textit{Options:}

\noindent\textit{A. [content for option A]}

\noindent\textit{B. [content for option B]}

\noindent\textit{C. [content for option C]}

\noindent\textit{D. [content for option D]}

~\

\noindent\textit{Answer:}

\end{preperformancebox}
    \end{minipage}
\end{preouterbox}
\caption{Instructions for memorization extraction and model performance evaluation, respectively.}
  \label{fig:pre_exp_prompt}
\end{figure*}

\begin{table*}
\centering
\begin{tblr}{
  row{2} = {c},
  cell{1}{1} = {r=2}{},
  cell{1}{2} = {r=2}{},
  cell{1}{3} = {c=2}{c},
  cell{1}{5} = {c=2}{c},
  cell{3}{1} = {r=3}{},
  cell{3}{3} = {c},
  cell{3}{4} = {c},
  cell{3}{5} = {c},
  cell{3}{6} = {c},
  cell{4}{3} = {c},
  cell{4}{4} = {c},
  cell{4}{5} = {c},
  cell{4}{6} = {c},
  cell{5}{3} = {c},
  cell{5}{4} = {c},
  cell{5}{5} = {c},
  cell{5}{6} = {c},
  cell{6}{1} = {r=3}{},
  cell{6}{3} = {c},
  cell{6}{4} = {c},
  cell{6}{5} = {c},
  cell{6}{6} = {c},
  cell{7}{3} = {c},
  cell{7}{4} = {c},
  cell{7}{5} = {c},
  cell{7}{6} = {c},
  cell{8}{3} = {c},
  cell{8}{4} = {c},
  cell{8}{5} = {c},
  cell{8}{6} = {c},
  cell{9}{1} = {r=3}{},
  cell{9}{3} = {c},
  cell{9}{4} = {c},
  cell{9}{5} = {c},
  cell{9}{6} = {c},
  cell{10}{3} = {c},
  cell{10}{4} = {c},
  cell{10}{5} = {c},
  cell{10}{6} = {c},
  cell{11}{3} = {c},
  cell{11}{4} = {c},
  cell{11}{5} = {c},
  cell{11}{6} = {c},
  cell{12}{1} = {r=3}{},
  cell{12}{3} = {c},
  cell{12}{4} = {c},
  cell{12}{5} = {c},
  cell{12}{6} = {c},
  cell{13}{3} = {c},
  cell{13}{4} = {c},
  cell{13}{5} = {c},
  cell{13}{6} = {c},
  cell{14}{3} = {c},
  cell{14}{4} = {c},
  cell{14}{5} = {c},
  cell{14}{6} = {c},
  cell{15}{1} = {r=3}{},
  cell{15}{3} = {c},
  cell{15}{4} = {c},
  cell{15}{5} = {c},
  cell{15}{6} = {c},
  cell{16}{3} = {c},
  cell{16}{4} = {c},
  cell{16}{5} = {c},
  cell{16}{6} = {c},
  cell{17}{3} = {c},
  cell{17}{4} = {c},
  cell{17}{5} = {c},
  cell{17}{6} = {c},
  cell{18}{1} = {r=3}{},
  cell{18}{3} = {c},
  cell{18}{4} = {c},
  cell{18}{5} = {c},
  cell{18}{6} = {c},
  cell{19}{3} = {c},
  cell{19}{4} = {c},
  cell{19}{5} = {c},
  cell{19}{6} = {c},
  cell{20}{3} = {c},
  cell{20}{4} = {c},
  cell{20}{5} = {c},
  cell{20}{6} = {c},
  hline{1,3,21} = {-}{},
  hline{2} = {3-6}{},
}
LLMs         & Dataset & $2 \times 2$ squares &                     & $3 \times 3$ squres &                     \\
             &         & Ratio (\%)                        & Pearson correlation & Ratio (\%)                       & Pearson correlation \\
LlaMA2-Qwen  & All     & 38.57                             & -0.7755             & 74.17                            & -0.8124             \\
             & Simple  & 37.07                             & -0.7784             & 72.63                            & -0.8121             \\
             & Pro     & 38.66                             & -0.783              & 74.51                            & -0.8109             \\
LlaMA2-GPT  & All     & 35.22                             & -0.7835             & 71.04                            & -0.7924             \\
             & Simple  & 33.9                              & -0.777              & 69.62                            & -0.7919             \\
             & Pro     & 35.45                             & -0.7916             & 71.54                            & -0.7881             \\
Mistral-Qwen & All     & 44.9                              & -0.8722             & 80.82                            & -0.8794             \\
             & Simple  & 38.47                             & -0.8494             & 74.04                            & -0.8271             \\
             & Pro     & 44.32                             & -0.8045             & 80.08                            & -0.8682             \\
Mistral-GPT  & All     & 40.37                             & -0.8042             & 76.58                            & -0.8736             \\
             & Simple  & 35.51                             & -0.8297             & 72.27                            & -0.8664             \\
             & Pro     & 38.52                             & -0.7103             & 74.91                            & -0.7969             \\
Vicuna-Qwen  & All     & 42.94                             & -0.8771             & 79.23                            & -0.8365             \\
             & Simple  & 37.86                             & -0.758              & 73.85                            & -0.7168             \\
             & Pro     & 42.01                             & -0.8609             & 77.86                            & -0.886              \\
Vicuna-GPT   & All     & 38.69                             & -0.8621             & 74.83                            & -0.8672             \\
             & Simple  & 34.77                             & -0.8096             & 70.71                            & -0.7775             \\
             & Pro     & 37.37                             & -0.7794             & 73.98                            & -0.8728             
\end{tblr}
\caption{The ratio and the Pearson-correlation between the $F_c$ and $F_m$ of the MCQs within the upper right and lower left $2 \times 2$ and $3 \times 3$ squares. For LLMs, `LlaMA2-Qwen' refers that the $F_c$ and $F_m$ are calculated with LlaMA2 based on the Qwen-extracted triplet, and similarly hereinafter. For the Dataset column, `All' stands for all the qualified MCQs after the triplet extraction, `Pro' refers to the qualified MCQs that are the members of the mmlupro dataset while `Simple' refers to the rest of the MCQs that are relatively easier.}\label{tab:mem_cap}
\end{table*}

\begin{figure*}[ht]
\centering
\begin{tcolorbox}[title={Prompt template for triplet extraction}, width=14cm]
\it\footnotesize{You are an expert of Knowledge Keyword extraction. Analyze and summarize the Question based on the given Fact corpus and extract the Knowledge Keyword, the Attribute and the Context (if necessary) within the Question.

~

Given a Fact corpus, a Question about the Fact corpus, and the Answer to the Question, analyze the Question corpus as well as the given Answer. Applying the provided steps, extract the Knowledge Keyword, the Attribute of the Knowledge Keyword and the necessary Context to obtain the key information of the Question, ensuring they are sufficient for answering the given Question and obtaining the given Answer.

~

\# Steps

~

1. **Review the Fact corpus:** Read through the entire Fact corpus to understand the context.

~

2. **Identify the Question:** Focus on the given Question to capture which part of the Fact corpus it is asking about.

~

3. **Understand the Answer to the Question:** Compare the given Answer and the identified questioned part within the Fact corpus and understand why this answer was chosen.

~

4. **Write Step-by-Step Reasoning:**

- Identify the asked Knowledge Keyword in the Question that is the subject of the most information in the Fact corpus and the asked Question is about the information among.

- Determine the asked Attribute of the Knowledge Keyword in the Question, which can be used to infer the given Answer.

- Review the identified Knowledge Keyword and Attribute to confirm that only these two parts can be used to obtain the given Answer to the given Question. If not, extract all the necessary Context from the Question that makes it enough to obtain the given Answer to the given Question.

~

5. **Determine Outcome:** Based on the reasoning, conclude and extract the Knowledge Keyword, the Attribute and the Context (if necessary) of the Question according to the Question corpus.

~

\# Output Format

~

Provide the outcome in the following format:

~

- **Step-by-Step Reasoning:** [Detailed reasoning here]

- **Knowledge Keyword:** [Extracted Knowledge Keyword here]

- **Attribute:** [Extracted Attribute of the Knowledge Keyword here]

- **Context:** [Extracted Context within the Question to make up for the Knowledge Keyword and the Attribute here if necessary]

~

\# Examples

~

[examples]

~

\# Notes

~

- Strictly follow the format of the examples and give Knowledge Keywords, the Attribute and the Context (if necessary) anyway.

- The extracted Knowledge Keyword, Attribute and Context (if necessary) should be the original text within the Question and should not incorporate any phrases that cannot be exactly matched in the Question.

- Never include any information from the options of the multiple choice question, especially the content of the answer option.

- The extracted Knowledge Keyword, Attribute and Context (if necessary) should include all the necessary information only within the Question Corpus for answering the Question and obtaining the given Answer.

~

**Fact:** [question]\quad[option content list]\quad[subject]\quad[answer option index][answer option ID]

~

**Question:** [question]

~

**Answer:** [content of the answer option]}

\end{tcolorbox}

\caption{Instruction for LLMs to extract the triplet.}
\label{fig:reorg}
\end{figure*}

\begin{figure*}[ht]
\centering
\begin{tcolorbox}[title={Prompt template for triplet validation \& reflection}, width=14cm]
\it\footnotesize{You are an expert of [subject] and an advanced reasoning agent that can determine whether the given Knowledge Keyword, Attribute of the Knowledge Keyword and the Context present most of the necessary information of the Question for obtaining the given Answer. Suppose you have sufficient background knowledge about {subj}. Consider the given Knowledge Keyword, Attribute and the Context, then determine whether the given Answer can be directly obtained from them even without the Question.

~

\# Steps

~

1. **Check the Semantic completeness:** Suppose you have sufficient background knowledge about [subject], and you can solve the given Question and obtain the given Answer. Read through the given Knowledge Keyword, Attribute, Context and the given Question. Check if the given Knowledge Keyword, Attribute, Context are the original text within the Question and contain the necessary queried information the Question itself provided (ignore the information the Question did not provided). If not so, check if the missed information is indeed incorporated in the Question (which is not acceptable, but if not, it is acceptable). Point out the information that is within the Question but they have missed. Then in a few sentences, diagnose the possible reason for failure or the phrasing discrepancy, and devise new, concise, high-level improvement suggestions to avoid the same failure.

~

2. **Check the Answer relevance:** Suppose you have sufficient background knowledge about {subj}, and you can solve the given Question and obtain the given Answer. Read through the given Knowledge Keyword, Attribute, Context and the given Question. Read through the given Knowledge Keyword, Attribute, Context and the given Answer. Check if the Answer can be directly inferred with the given Knowledge Keyword, Attribute and the Context without seeing the Question. If not so, check if the missed information is indeed incorporated in the Question (which is not acceptable, but if not, it is acceptable). Point out the information that is within the Question but they have missed. Then in a few sentences, diagnose the possible reason for failure or the phrasing discrepancy, and devise new, concise, high-level improvement suggestions to avoid the same failure.

~

3. **Check the Semantic Redundancy:** Read through the given Knowledge Keyword, Attribute, Context, the given Question and the given corresponding Answer. Check if the Answer can be directly matched within the given Knowledge Keyword, Attribute and the Context. Check if there are any unnecessary information within the given Knowledge Keyword, Attribute and the Context for obtaining the given Answer to the Question. If not so, point out what is redundant. Then in a few sentences, diagnose the possible reason for failure or the phrasing discrepancy, and devise new, concise, high-level improvement suggestions to avoid the same failure.

~

\# Output Format

~

Provide the outcome in the following format:

~

- **Step-by-Step Reasoning:** [Detailed reasoning here]

- **Verdict for the given Knowledge Keyword, Attribute and Context:** [Single verdict (Yes/No) here for whether the given Knowledge Keyword, Attribute and Context contain most of the asked information of the Question, can be used to infer the given Answer with only them without the whole Question, and do not contain redundant information for obtaining the given Answer.]

~

\# Notes

~

- Do not deviate from the specified format. Do not generate anything else after the Verdict (only Yes/No) for the given Knowledge Keyword, Attribute and Context.

- Suppose you have sufficient background knowledge about {subj}, and you can solve the given Question and obtain the given Answer. For Semantic completeness and Answer relevance, it is acceptable to miss information that is also not incorporated in the Question.

- Provide a detailed explanation following the given steps before arriving at the verdict (Yes/No). Provide a final verdict (only Yes/No) in order at the end in the given format.

~

- **Question:** [question]

- **Answer:** [answer]

~

- **Knowledge Keyword:** [extracted knowledge entity]

- **Attribute:** [extracted attribute]

- **Context:** [extracted context]}

\end{tcolorbox}

\caption{Instruction for LLMs to verify whether the triplet is qualified and reflect.}
\label{fig:reorg_verdict}
\end{figure*}

\begin{figure*}[ht]
\centering
\begin{tcolorbox}[title={Prompt template for the second round triplet extraction}, width=18cm]
\it\scriptsize{You are an advanced reasoning agent that can improve through self-reflection and an expert of Knowledge Keyword extraction. Analyze and summarize the Question based on the given Fact corpus and extract the Knowledge Keyword, the Attribute and the Context (if necessary) within the Question.

~

Given a Fact corpus, a Question about the Fact corpus, and the Answer to the Question, analyze the Question corpus as well as the given Answer. Applying the provided steps, extract the Knowledge Keyword, the Attribute of the Knowledge Keyword and the necessary Context to rephrase the Question, ensuring they are sufficient for answering the given Question and obtaining the given Answer.

~

\# Steps

~

1. **Review the Fact corpus:** Read through the entire Fact corpus to understand the context.

~

2. **Identify the Question:** Focus on the given Question to capture which part of the Fact corpus it is asking about.

~

3. **Understand the Answer to the Question:** Compare the given Answer and the identified questioned part within the Fact corpus and understand why this answer was chosen.

~

4. **Write Step-by-Step Reasoning:**

- Identify the asked Knowledge Keyword in the Question that is the subject of the most information in the Fact corpus and the asked Question is about the information among.

- Determine the asked Attribute of the Knowledge Keyword in the Question, which can be used to infer the given Answer.

- Review the identified Knowledge Keyword and Attribute to confirm that only these two parts can be used to obtain the given Answer to the given Question. If not, extract all the necessary Context from the Question that makes it enough to obtain the given Answer to the given Question.

~

5. **Determine Outcome:** Based on the reasoning, conclude and extract the Knowledge Keyword, the Attribute and the Context (if necessary) of the Question according to the Question corpus.

~

\# Output Format

~

Provide the outcome in the following format:

~

- **Step-by-Step Reasoning:** [Detailed reasoning here]

- **Knowledge Keyword:** [Extracted Knowledge Keyword here]

- **Attribute:** [Extracted Attribute of the Knowledge Keyword here]

- **Context:** [Extracted Context within the Question to make up for the Knowledge Keyword and the Attribute here if necessary]

~

\# Examples

~

[examples]

~

You will be given a previous trial. You were unsuccessful in extracting the Knowledge Keyword, Attribute and the necessary that meet the requirements in the previous trial. Given the Reflection below, improve the process. The process is as follows:

~

\# Previous returns:

~

- **Fact:** [question]\quad[option content list]\quad[subject]\quad[answer option index][answer option ID]

~

- **Question:** [question]

~

- **Answer:** [answer option content]

~

- **Knowledge Keyword:** [extracted knowledge entity of the last trial]

~

- **Attribute:** [attribute of the last trial]

~

- **Context:** [context of the last trial]

~

- **Reflection:**

[rational of the last trial]

~

\# Notes

~

- Consider the Reflection given above. Improve the extraction of Knowledge Keyword, Attribute and Context (if necessary).

- Strictly follow the format of the examples and give Knowledge Keywords, the Attribute and the Context (if necessary) anyway.

- The extracted Knowledge Keyword should be phrases within the Question and should not incorporate any information of the Fact corpus or the given Answer that is not mentioned in the Question.

- The extracted Attribute and Context (if necessary) should only include information from the Question corpus. Never include information from the options of the multiple choice question, especially the content of the answer option.

- The extracted Knowledge Keyword, Attribute and Context (if necessary) should include all the necessary information only within the Question Corpus for answering the Question and obtaining the given Answer.

~

**Fact:** [question]\quad[option content list]\quad[subject]\quad[answer option index][answer option ID]

~

**Question:** [question]

~

**Answer:** [content of the answer option]}

\end{tcolorbox}

\caption{Instruction for LLMs on the second round triplet extraction.}
\label{fig:reorg_reflect}
\end{figure*}

\begin{figure*}[htbp]
\centering
\begin{outerbox}
    \noindent
    \begin{minipage}[t]{0.48\textwidth}
        \begin{otemplatebox}
            \it\footnotesize{You are an expert on multiple choice questions of [subject]. Analyze the given question and the given options. Determine the correct answer option to the question. 

~

Given a Question and the potential Answer options to the Question, analyze the Question as well as the given options. Generate the option ID of the correct option (answer).

~

- **Question:**

[question]

~

- **Options:**

A. [option A]

B. [option B]

C. [option C]

D. [option D]}
        \end{otemplatebox}
    \end{minipage}
    \hfill
    \begin{minipage}[t]{0.48\textwidth}
        \begin{oexamplebox}
            \it\footnotesize{You are an expert on multiple choice questions of high school computer science. Analyze the given question and the given options. Determine the correct answer option to the question. 

~

Given a Question and the potential Answer options to the Question, analyze the Question as well as the given options. Generate the option ID of the correct option (answer).

~

- **Question:**

Which of the following is usually NOT represented in a subroutine's activation record frame for a stack-based programming language?

~

- **Options:**

A. Values of local variables

B. A heap area

C. The return address

D. Stack pointer for the calling activation record}
        \end{oexamplebox}
    \end{minipage}
    
    \vspace{8pt}
    
    \noindent
    \begin{minipage}[t]{0.48\textwidth}
        \begin{rtemplatebox}
            \it\footnotesize{You are an expert on multiple choice questions of [subject]. Analyze the given Knowledge Entity, Attribute of the Knowledge Entity, the Context of a question, and the given options to the question. Determine the correct answer option to the question.

~

The Knowledge Entity is the questioned subject of the question. The Attribute is the questioned attribute of the Knowledge Entity, and the Context is the necessary context information for answering the question. Given a set of Knowledge Entity, Attribute, and Context (which three are extracted as the key information from a question), and the potential Answer options to the Question, analyze the given Knowledge Entity, Attribute, Context as well as the options. Generate the option ID of the correct option (answer).

~

- **Knowledge Entity:**

[knwoledge entity]

~

- **Attribute:**

[attribute]

~

- **Context:**

[context]

~

- **Options:**

A. [option A]

B. [option B]

C. [option C]

D. [option D]}
        \end{rtemplatebox}
    \end{minipage}
    \hfill
    \begin{minipage}[t]{0.48\textwidth}
        \begin{rexamplebox}
            \it\footnotesize{You are an expert on multiple choice questions of high school computer science. Analyze the given Knowledge Entity, Attribute of the Knowledge Entity, the Context of a question, and the given options to the question. Determine the correct answer option to the question.

~

The Knowledge Entity is the questioned subject of the question. The Attribute is the questioned attribute of the Knowledge Entity, and the Context is the necessary context information for answering the question. Given a set of Knowledge Entity, Attribute, and Context (which three are extracted as the key information from a question), and the potential Answer options to the Question, analyze the given Knowledge Entity, Attribute, Context as well as the options. Generate the option ID of the correct option (answer).

~

- **Knowledge Entity:**

subroutine's activation record frame

~

- **Attribute:**

usually NOT represented

~

- **Context:**

for a stack-based programming language

~

- **Options:**

A. Values of local variables

B. A heap area

C. The return address

D. Stack pointer for the calling activation record}
        \end{rexamplebox}
    \end{minipage}
\end{outerbox}
\caption{Template and an example of the Original MCQ template and the \modelname MCQ template. [$\cdot$] refers to the blank that should be filled according to the content of each MCQ.}
\label{fig:mcq_form}
\end{figure*}

\begin{figure*}[ht]
  \centering
  \subfloat[Probability and p-value with Llama2 based on GPT.\label{fig:conti_subfig1}]{%
    \includegraphics[width=0.325\textwidth]{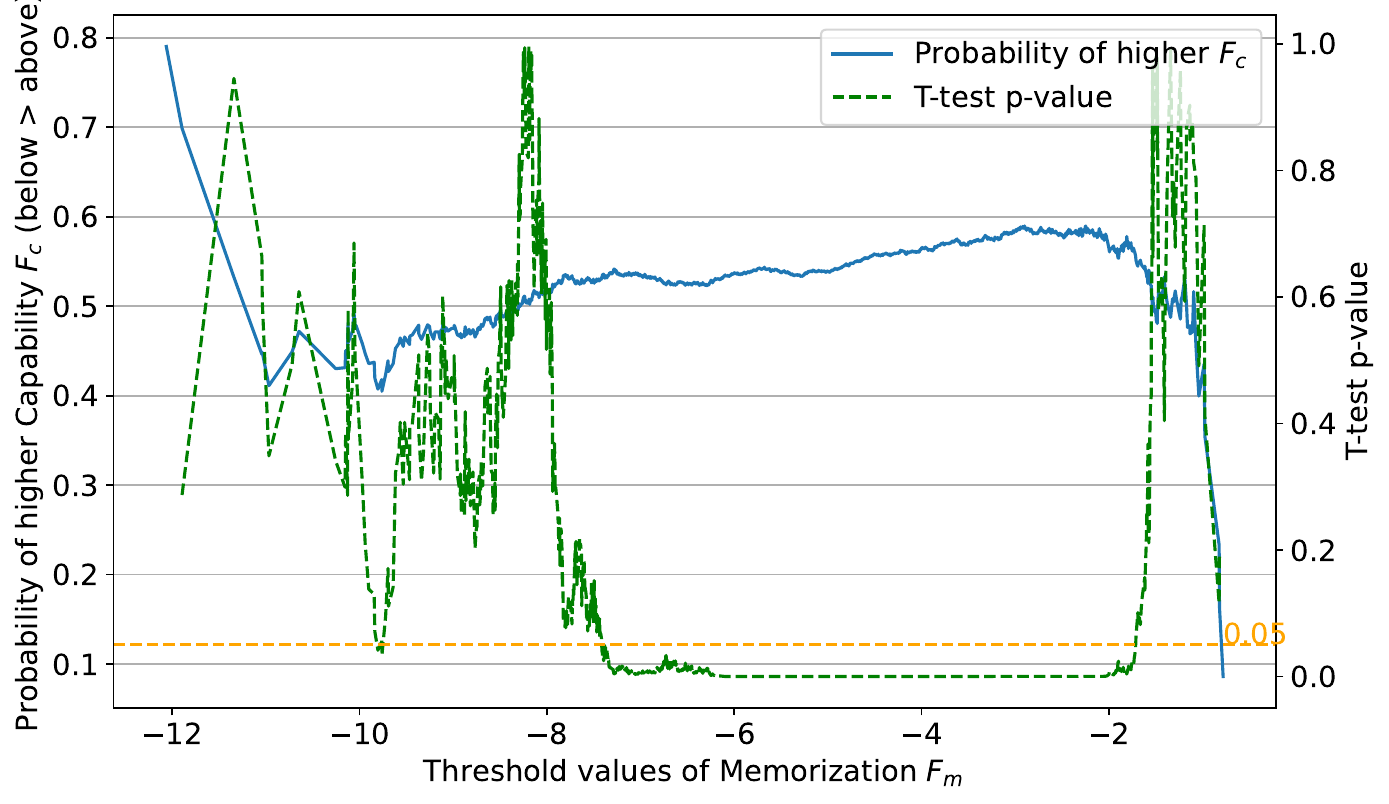}%
  }\hfill
  \subfloat[Probability and p-value with Mistral based on GPT.\label{fig:conti_subfig2}]{%
    \includegraphics[width=0.325\textwidth]{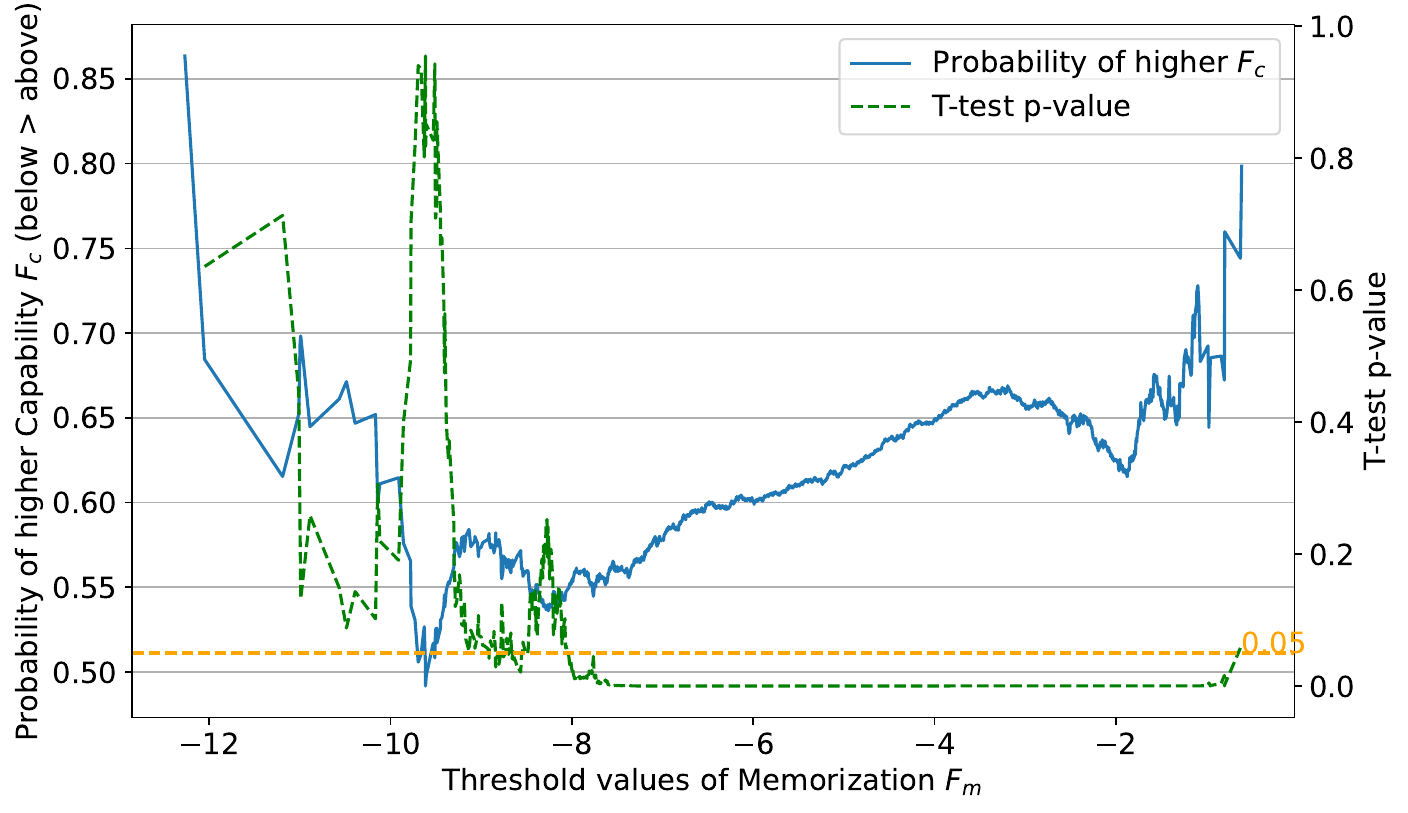}%
  }\hfill
  \subfloat[Probability and p-value with Vicuna based on GPT.\label{fig:conti_subfig3}]{%
    \includegraphics[width=0.325\textwidth]{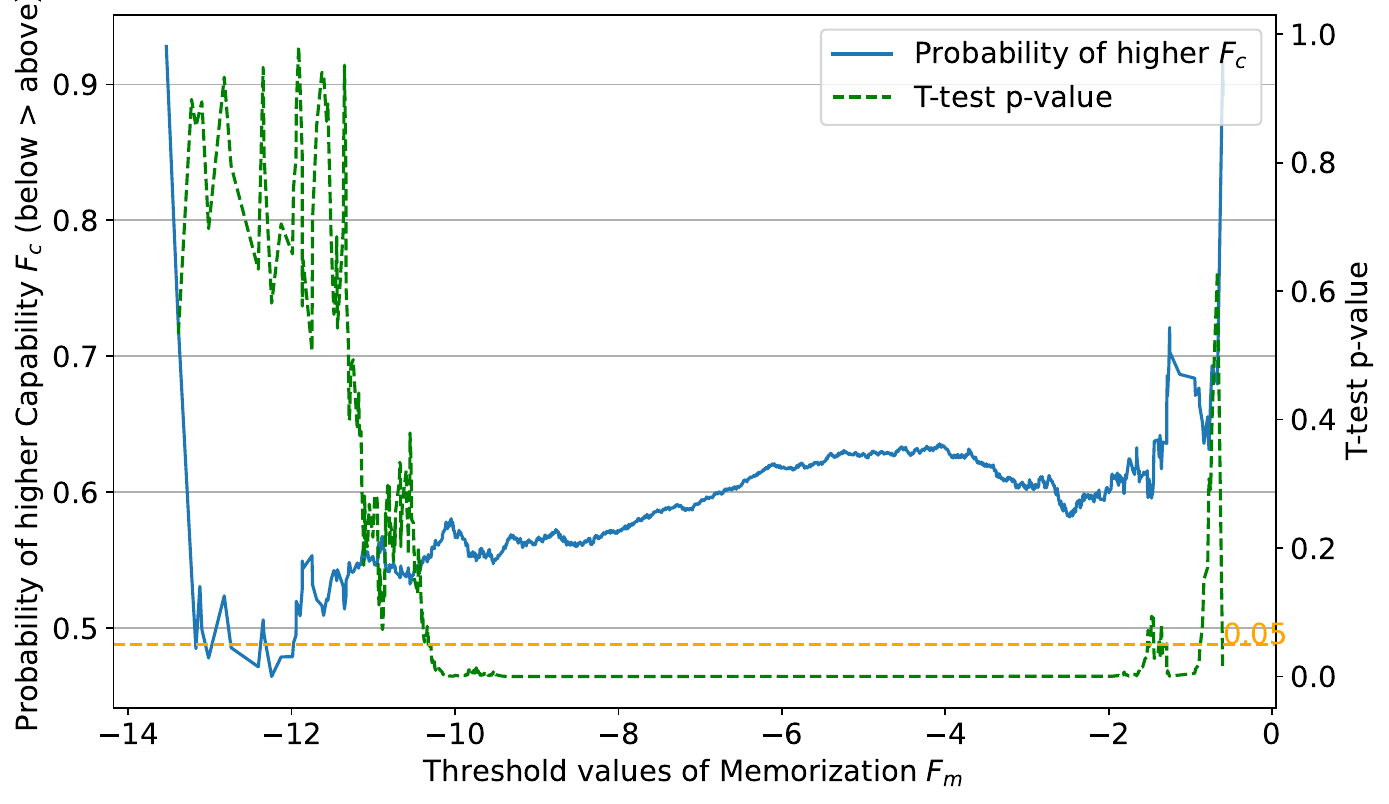}%
  }

  \vfill  

  \subfloat[Probability and p-value with Llama2 based on GPT (21.59\,\% MCQs filtered).\label{fig:conti_subfig4}]{%
    \includegraphics[width=0.325\textwidth]{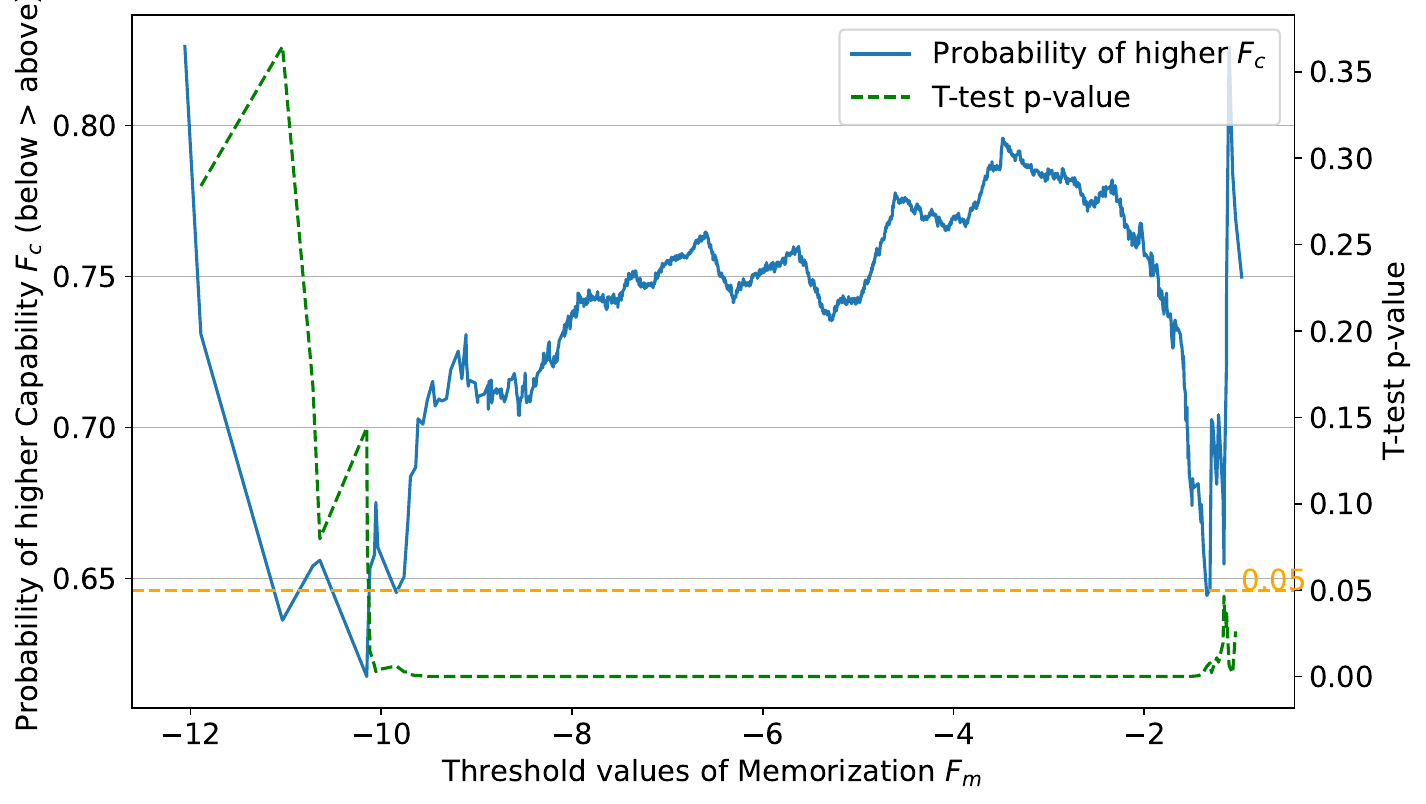}%
  }\hfill
  \subfloat[Probability and p-value with Mistral based on GPT (15.95\,\% MCQs filtered).\label{fig:conti_subfig5}]{%
    \includegraphics[width=0.325\textwidth]{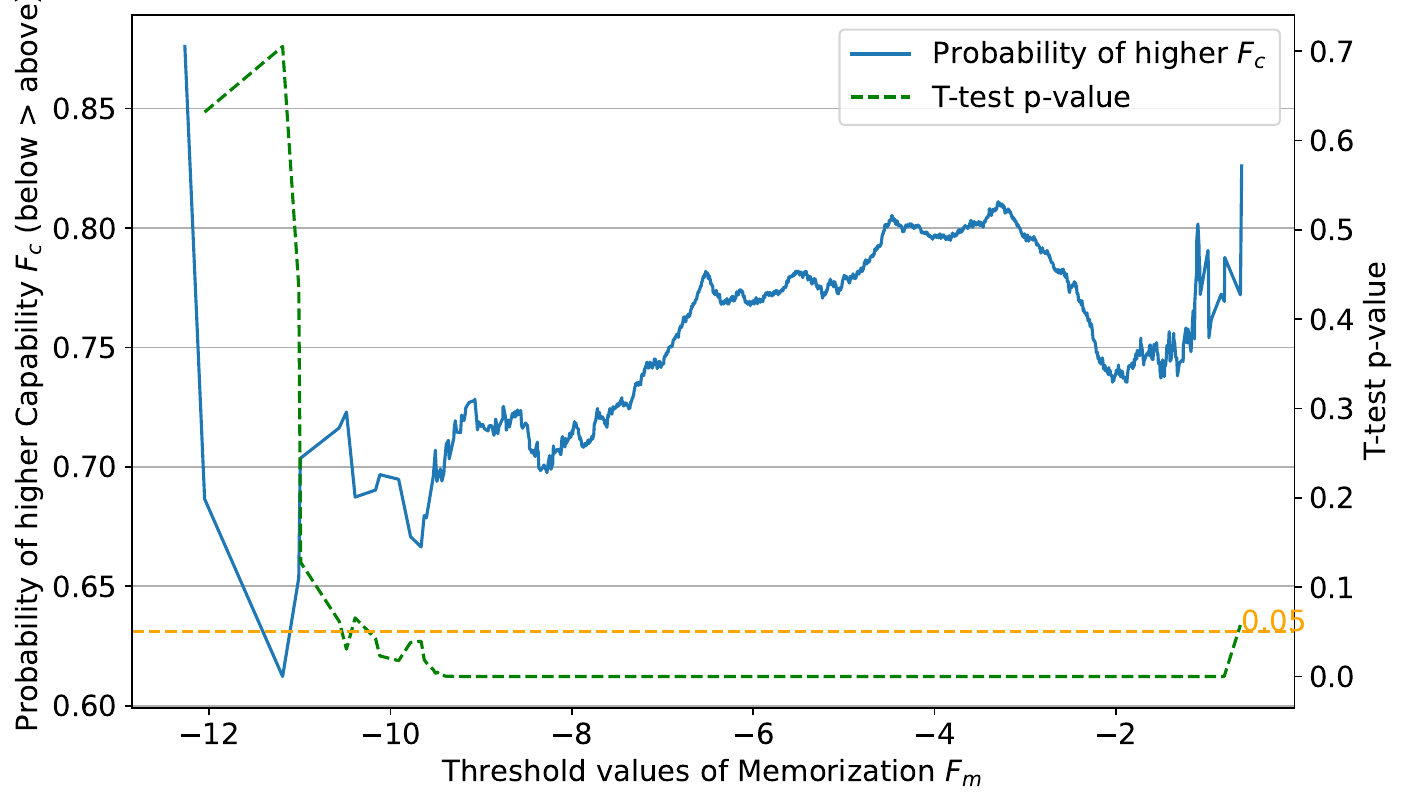}%
  }\hfill
  \subfloat[Probability and p-value with Vicuna based on GPT (17.59\,\% MCQs filtered).\label{fig:conti_subfig6}]{%
    \includegraphics[width=0.325\textwidth]{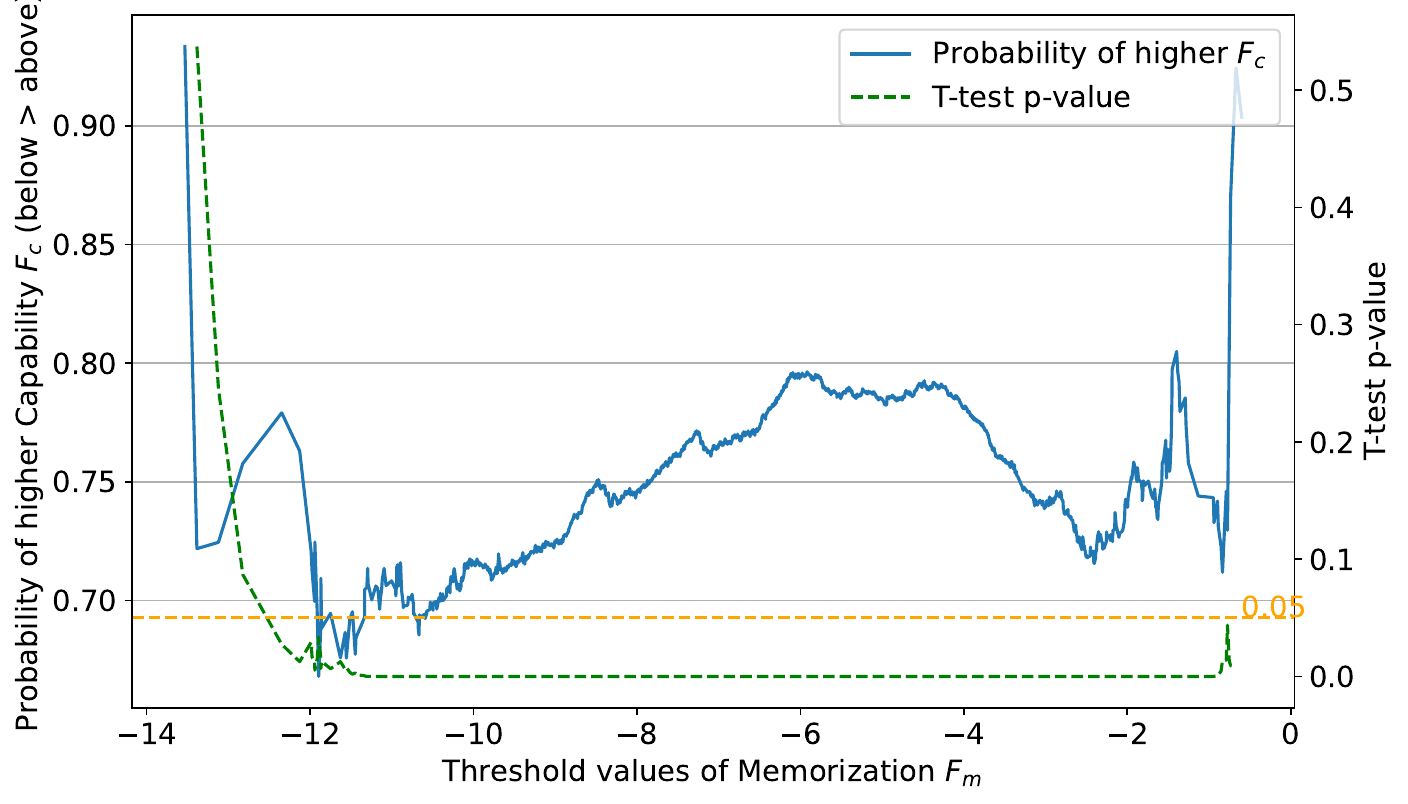}%
  }

  \caption{%
    The over-performing probability curve and p-value curve with different $F_m$ thresholds. In this figure, we take each unique $F_m$ as the threshold to separate the qualified MCQs into the Memorized and non-Memorized MCQs. We compute the probability of a randomly selected non-Memorized MCQ's $F_c$ exceeding a randomly selected Memorized MCQ's $F_c$ under each threshold as the blue curve, and the green curve is the p-value of the T-test between the $F_c$s of the non-Memorized MCQs and the Memorized MCQs.
  }
  \label{fig:conti_mem_cap}
\end{figure*}

}


 
%

\newcommand{\BIBentryALTinterwordspacing}{}
\newcommand{\BIBentrySTDinterwordspacing}{}

\newpage

\end{document}